\pdfoutput=1
\documentclass[twoside]{article}

\usepackage[preprint]{aistats2027}
\renewcommand{\acknowledgments}{\ackaccepted}
\expandafter\ifx\csname iflatexml\endcsname\relax
\expandafter\newif\csname iflatexml\endcsname
\fi

\usepackage[round]{natbib}

\usepackage[utf8]{inputenc}
\usepackage[T1]{fontenc}
\usepackage{url}
\usepackage{booktabs}
\usepackage{nicefrac}
\usepackage{microtype}
\usepackage{xcolor}

\usepackage{amsmath}
\usepackage{amssymb}
\usepackage{amsthm}
\usepackage{mathtools}
\usepackage{dsfont}  %

\usepackage{enumitem}

\usepackage{float}  %
\usepackage{algorithm}
\usepackage[noend]{algpseudocode}  %

\usepackage{graphicx}

\usepackage{hyperref}

\newtheorem{theorem}{Theorem}[section]
\newtheorem{lemma}[theorem]{Lemma}
\newtheorem{corollary}[theorem]{Corollary}
\newtheorem{proposition}[theorem]{Proposition}
\newtheorem{definition}[theorem]{Definition}
\newtheorem{remark}[theorem]{Remark}

\newcommand{\R}{\mathbb{R}}
\newcommand{\E}{\mathbb{E}}
\newcommand{\Prob}{\mathbb{P}}
\newcommand{\indicator}{\mathds{1}}
\newcommand{\eps}{\varepsilon}
\newcommand{\cD}{\mathcal{D}}
\newcommand{\cA}{\mathcal{A}}
\newcommand{\cZ}{\mathcal{Z}}
\newcommand{\thetahat}{\hat{\theta}}

\newcommand{\papertitle}{A Theoretical Reduction From Membership Inference to Reconstruction Attacks Against Learners}

\newcommand{\paperauthors}{%
  Amnon Levi \And Yuval Dagan%
}
\newcommand{\paperaddresses}{%
  School of Computer Science and AI \\
  Tel Aviv University \\
  Tel Aviv, Israel \\
  \texttt{amnonlevi@mail.tau.ac.il}
  \And
  School of Computer Science and AI \\
  Tel Aviv University \\
  Tel Aviv, Israel \\
  \texttt{ydagan@tauex.tau.ac.il}%
}

\begin{document}

\iflatexml
\title{\papertitle}
\author{Amnon Levi \\ {School of Computer Science and AI} \\ {Tel Aviv University} \\ {Tel Aviv, Israel} \\ \texttt{amnonlevi@mail.tau.ac.il}
\and
Yuval Dagan \\ {School of Computer Science and AI} \\ {Tel Aviv University} \\ {Tel Aviv, Israel} \\ \texttt{ydagan@tauex.tau.ac.il}}
\maketitle
\else
\twocolumn[

\aistatstitle{\papertitle}

\aistatsauthor{\paperauthors}

\aistatsaddress{\paperaddresses} ]
\fi

\begin{abstract}
We study \emph{reconstruction attacks}, which recover training data from a learned model. In our adversarial model, the adversary designs a hard instance, drawn from a parameterized family of data distributions, and is given the parameter used to generate it. Our main contribution is a general reduction showing that, in this model, \emph{membership inference} (determining whether a given datapoint was used for training) implies reconstruction: for a training set of size $n$, a membership test with recall at least $q(n)$ and false-positive rate $O(\delta^2/n^2)$ yields a reconstruction algorithm that, with probability $1 - O(\delta)$, outputs only training points, and recovers $\Omega(n \cdot q(n))$ of them in expectation. This reduction applies to any learning algorithm.
For stochastic convex optimization (SCO), we make the reduction \emph{computationally efficient} over exponentially large domains via structured distributions whose support is partitioned into polynomially many categories of exponentially many elements each, so that reconstruction iterates over categories rather than the full support. Combining this with an existing membership inference result for SCO, we prove that any learning algorithm achieving low excess risk over all distributions admits a polynomial-time reconstruction attack recovering $\Omega(n)$ training points, even when the underlying distribution has exponentially large support.
\end{abstract}

\section{Introduction}
\label{sec:intro}

Despite their widespread adoption, machine learning models remain vulnerable to a range of information retrieval attacks, exploits that leak private information through model parameters. These threats vary in what the adversary aims to learn:
\begin{itemize}[itemsep=1pt, topsep=2pt]
    \item \emph{Property inference}: inferring global statistical properties of the training data that the model producer did not intend to share, such as demographic composition or environmental conditions \citep{ateniese2015hacking, ganju2018property}.
    \item \emph{Membership inference}: determining whether a specific datapoint was part of the training data \citep{shokri2017membership, yeom2018privacy}.
    \item \emph{Model inversion}: exploiting model access to infer sensitive training data features, ranging from recovering representative inputs for a class to inferring unknown attributes of known individuals \citep{fredrikson2014privacy, fredrikson2015model}.
    \item \emph{Reconstruction}: directly extracting actual training data from the model. Such attacks can expose sensitive personal information by, for example, extracting verbatim text including names, phone numbers, and email addresses from a language model, or reconstructing training images from neural network classifiers \citep{carlini2019secret,carlini2021extracting, balle2022reconstructing, haim2022reconstructing}.
\end{itemize}

The ability to extract training data implies that models ``memorize'' portions of their training data. One might hope to mitigate this risk by avoiding memorization entirely. However, \citet{feldman2020does} established that for long-tailed distributions, which are characteristic of natural data, memorization of individual labels, including noisy and outlier labels, is \emph{necessary} for achieving close-to-optimal generalization error. This creates a fundamental tension: the very property that enables learning may also enable information retrieval attacks.

Differential privacy (DP) offers one approach to this problem. DP schemes provide mathematically proven guarantees on information leakage by adding noise during training \citep{dwork2006calibrating, abadi2016deep}. In principle, DP can reduce leaked information to arbitrarily small levels, making it the primary formal defense against the privacy vulnerabilities described above. However, achieving meaningful DP guarantees comes at a significant cost to model utility: the noise required for strong privacy typically degrades accuracy \citep{carlini2021extracting}, and the assumptions underlying DP -- that private information maps neatly to individual training records -- may not hold for natural language and other complex data \citep{brown2022language}.

This raises a fundamental question: differential privacy greatly restricts information leakage, but is such a strong guarantee necessary to prevent reconstruction?

To address this question rigorously, we begin by examining the relationship between membership inference and reconstruction. Membership inference determines whether a \emph{given} candidate belongs to the training set, while reconstruction must \emph{output} training data, knowing at most their distribution. Reconstruction is a stronger requirement: a reconstruction algorithm implicitly solves membership inference for every datapoint it outputs, but the converse fails in general; we show that it holds when the adversary designs the data distribution (Section~\ref{sec:reduction}).

Membership inference has received substantial theoretical attention. Theoretical bounds quantitatively relate the success of membership inference to overfitting \citep{yeom2018privacy}, and theoretically optimal attack strategies have been derived for various settings \citep{sablayrolles2019white, carlini2022membership}. By contrast, reconstruction attacks have been studied primarily empirically -- demonstrating extraction of training data from language models \citep{carlini2021extracting} -- or theoretically only for restricted model classes \citep{haim2022reconstructing, balle2022reconstructing, voitovych2025traceability, cohen2024reconstruction}. General theoretical bounds for reconstruction, however, have been largely absent.

This gap between membership inference theory and reconstruction practice motivates our central question: \emph{Can membership inference guarantees be systematically converted into reconstruction guarantees?} If so, the rich theoretical toolkit for analyzing membership inference could be leveraged to establish reconstruction bounds.

We study the bridging of this gap in both a general context and in the \emph{stochastic convex optimization} (SCO) setting. SCO provides an ideal setting for our investigation for two reasons. First, it underlies gradient-based optimization, the workhorse behind modern machine learning \citep{shalev2009stochastic}. Second, the gap between private and non-private learning is sharp and well-quantified: without privacy constraints, $\Theta(1/\eps^2)$ samples suffice for learning \citep{agarwal2012information, feldman2016generalization}, while DP-compliant algorithms require sample complexity that grows with the dimension $d$ \citep{bassily2014private}, making SCO a natural setting to ask whether the full strength of DP is necessary to prevent reconstruction, or whether the privacy cost can be avoided.

Formally, in SCO a learner receives $n$ datapoints from a distribution $\cD$ over a domain $\cZ$ and outputs a parameter $\thetahat$ minimizing the population risk $F(\theta) = \E_{z \sim \cD}[f(\theta, z)]$, where both $f(\cdot, z)$ and $\Theta$ are convex. The goal is to achieve excess risk $F(\thetahat) - \min_\theta F(\theta) \le \eps$ with minimal sample complexity $n$.

In the privacy field under this setting, recent work by \citet{attias2024information} established an important impossibility result. Any \emph{$\eps$-learner}, an algorithm that achieves excess risk at most $\eps$ for \emph{all} distributions (i.e., with distribution-free guarantees), must necessarily ``memorize'' its training data in a precise sense. Specifically, for any $\eps$-learner there exists a (learner-dependent) hard distribution for which a membership inference test correctly identifies training datapoints with constant probability while maintaining a low false-positive rate. This result advances the membership inference research in SCO, and our approach aims to translate this advancement, as well as similar ones, to reconstruction.

The existence of a fundamental setting -- stochastic convex optimization -- where a membership test is inevitably highly probable raises two questions:
(i)~\emph{Is reconstruction also always highly probable in this setting?}
(ii)~\emph{Can it be done efficiently?}
Our aim is to answer these two questions and provide a framework for answering them for future settings as well.

\paragraph{Our contributions.} We take a step towards bridging the gap between membership inference and reconstruction with three main results, establishing a general theoretical framework for reconstruction attacks on learning algorithms.

\textbf{1. Reduction from Membership Inference to Reconstruction.} Our first contribution is framework-agnostic, and does not require the assumptions made in SCO. We show that if a setting admits a per-learner hard distribution for membership inference, then the same setting admits a per-learner hard distribution for reconstruction: the reconstruction-hard distribution is uniform over a candidate set sampled from the inference-hard distribution, provided the test keeps its guarantees under it (Definition~\ref{def:detection-rate}). Quantitatively, for a training set of size $n$, if the data distribution is uniform over a candidate set and a membership test has recall at least $q(n)$ (Definition~\ref{def:detection-rate}) and false-positive rate $O(\delta^2/n^2)$, then an algorithm that applies the test to every candidate outputs, with probability $1 - O(\delta)$, only training datapoints, and $\Omega(n \cdot q(n))$ of them in expectation. This reduction is \emph{learner-agnostic} and \emph{test-agnostic}: it applies to any learner and any membership test satisfying the stated conditions.

\textbf{2. Efficient Reconstruction on Exponential Domains.} \emph{A general construction.} The reduction in (1) enumerates over the candidate set, so its runtime scales with the size of the data domain, which is prohibitive when the domain is exponentially large. Our second contribution addresses this limitation: we give a construction that makes reconstruction computationally efficient even over exponentially large supports, by modeling the distribution as a polynomially large set of categories, each of which is an exponentially large set. We argue that this structure mirrors a feature found in many natural data distributions: a polynomial number of high-level categories, each populated by exponentially many fine-grained instances. For example, the set of all natural images is exponentially large, but it can be split into a (still large, but not exponential) set of trigram descriptions, e.g.,~``three golden triangles'', ``plumber eating spaghetti'', ``swirly purple fruit'', etc.

We construct a distribution that fits this model, for which the adversary produces a single candidate from each category set; with probability $1 - O(\delta)$, for a constant fraction of the training points, the candidate of the corresponding category coincides exactly with that training point. Such an adversary thus only needs a sufficiently differentiating inference test while iterating over the polynomial set of categories.

\textbf{3. Impossibility Theorem for SCO.} \emph{An application to SCO.} Combining the above, we establish a strong impossibility result: any $\eps$-learner (with distribution-free guarantees) is inherently vulnerable to reconstruction attacks. We obtain this by instantiating the category construction of (2) inside the SCO setting and establishing, for our own membership test and under the constructed block-uniform distribution, the two guarantees the reduction consumes: recall, carried over to it from the correlation bound of \citet[Lemma~6.3]{attias2024information} (Appendix~\ref{app:complete-proof}), and soundness (a low false-positive rate), which we prove directly, since their product-measure tail analysis does not apply to it (Lemma~\ref{lem:category-tail}, Corollary~\ref{cor:soundness}). The approach is modular: the category construction is setting-agnostic, and we instantiate it in SCO because that is, to our knowledge, the only learning setting with a membership inference impossibility at the rates the reduction requires; any future such result plugs into our reduction once its two rates hold under the hard distribution of Definition~\ref{def:game}.

Together, these results establish that for the linear SCO instance in the optimal sample complexity regime, any $\eps$-learner is vulnerable to reconstruction of $\Omega(n)$ training datapoints on some distribution of our family. Thus distribution-free accuracy precludes reconstruction resistance across distributions, a tension that goes beyond differential privacy.

\paragraph{Organization.} Section~\ref{sec:reduction} presents our reduction; Section~\ref{sec:exponential} develops the exponential-domain distribution construction and proves the main SCO reconstruction result; Section~\ref{sec:related} surveys related work. Full proofs and supporting lemmas appear in Appendices~\ref{app:algorithms}--\ref{app:technical} (notation is summarized in Appendix~\ref{app:technical}), and Appendix~\ref{app:discussion} discusses generalizations, necessary features of the adversarial model, and open directions.

\section{Reduction from Membership Inference to Reconstruction}
\label{sec:reduction}

In this section, we establish our first main result: a general reduction converting membership inference capabilities into reconstruction attacks. This reduction is learner-agnostic and applies to any setting with a membership test that is sufficiently accurate under the hard distribution of Definition~\ref{def:game}.

\subsection{Problem Setup}

We consider a general learning setting where a \emph{learning algorithm} $\cA = \{\cA_n\}_{n \in \mathbb{N}}$ is given as a sequence of maps $\cA_n: \cZ^n \to \Theta$, one for every sample size: on a training set $S_n = \{z_1, \ldots, z_n\}$ of datapoints drawn from a distribution over a domain $\cZ$, it outputs a model $\thetahat = \cA_n(S_n) \in \Theta$ (uppercase $Z_i$ denotes a random datapoint, lowercase $z$ a realized one). In stochastic convex optimization (Section~\ref{sec:exponential}) the learning problem is a \emph{convex problem instance}: a triple $(\Theta, \cZ, f)$ with $\Theta$ a convex subset of Euclidean space and the loss $f(\cdot, z)$ convex for every $z \in \cZ$; the reduction of this section requires no convexity.

\begin{definition}[Membership inference test]
\label{def:mi-test}
A \emph{membership inference test} is a function $T: \Theta \times \Theta \times \cZ \to \{0, 1\}$ that attempts to determine whether a datapoint $x \in \cZ$ was part of the training set, given the learned parameter $\thetahat$, the parameter $\theta$ of the data-generating process, and the datapoint $x$.
\end{definition}

\begin{definition}[Adversarial reconstruction game]
\label{def:game}
Fix a sample size $n$, a candidate-set size $N$, and a confidence parameter $\delta > 0$.
\begin{enumerate}[itemsep=1pt, topsep=2pt]
    \item The adversary designs a parameterized family of distributions $\{\cD_\theta\}_{\theta \in \Theta}$ over the domain $\cZ$ and a prior $\mathbf{D}$ over the parameters $\theta \in \Theta$.
    \item Nature draws $\theta \sim \mathbf{D}$. The adversary, knowing $\theta$, samples a candidate set $\Sigma = \{\sigma_1, \ldots, \sigma_N\} \sim \cD_\theta^{\otimes N}$ and defines the \emph{hard distribution} $\Delta_\theta := \mathrm{Uniform}(\Sigma)$.
    \item A training set $S \sim \Delta_\theta^{\otimes n}$ is drawn; the learner receives $S$ only, and outputs $\thetahat = \cA_n(S)$.
    \item The adversary receives $(\thetahat, \theta)$ and applies the membership test $T$ to the candidates in $\Sigma$.
\end{enumerate}
\end{definition}

The adversary thus constructs the family, the prior, and the hard distribution (steps 1--2) and is handed $\theta$ (step 4), so it has nothing to estimate. Some per-learner choice of hard distribution is \emph{necessary} at the standard sample size (Appendix~\ref{app:barrier}), while exact knowledge of $\theta$ relaxes to sampling access at polynomial cost in the SCO instantiation (Appendix~\ref{app:sampling}).

Theorem~\ref{thm:reduction} below assumes two guarantees on the membership test, defined next; both are required under the constructed $\Delta_\theta$, the distribution the learner is actually trained on, not under the ambient $\cD_\theta$. Section~\ref{sec:exponential} establishes both for the SCO instance under its own hard distribution $\Delta_\mu$: soundness in Section~\ref{sec:soundness}, and recall through the correlation bound (Lemma~\ref{lem:correlation}, proven for the product Bernoulli family $\cD_\mu$ and applied to $\Delta_\mu$ in Appendix~\ref{app:complete-proof}).

\begin{definition}[Recall and false-positive rates]
\label{def:detection-rate}
Let $T$ be a membership inference test, played in the game of Definition~\ref{def:game}. $T$ has \emph{recall} at least $q(n)$ and \emph{false-positive rate} at most $\beta$, with confidence $\delta$, if with probability $1 - \delta$ over $\theta \sim \mathbf{D}$ and the adversary's construction of $\Delta_\theta$ (the sampling of $\Sigma$), the following hold, where all inner probabilities and expectations are over $S \sim \Delta_\theta^{\otimes n}$ and any internal randomness of $\cA$ and $T$:
\begin{align}
&\E\Big[\big|\{x \in S : T(\cA_n(S), \theta, x) = 1\}\big|\Big] \ge q(n) \cdot n, \label{eq:true-positive}\\
&\Prob\big[T(\cA_n(S), \theta, x) = 1 \mid x \notin S\big] \le \beta \nonumber\\
&\qquad \text{for every candidate } x \in \Sigma. \label{eq:false-positive}
\end{align}
The conditioning in~\eqref{eq:false-positive} is well defined, as each candidate lands in $S$ with probability $\approx n/N$ under the finite-support $\Delta_\theta$. Condition~\eqref{eq:true-positive} says that the test flags, in expectation, at least a $q(n)$ fraction of the training points (its true-positive rate over the training set); this count form is what Step~5 of the proof of Theorem~\ref{thm:reduction} uses.
\end{definition}

\subsection{Main Reduction Theorem}

\begin{theorem}[Reduction from Membership Inference to Reconstruction]
\label{thm:reduction}
Consider the game of Definition~\ref{def:game} with $N = n^2/\delta$, played by a learning algorithm $\cA = \{\cA_n\}_{n \in \mathbb{N}}$, a family $\{\cD_\theta\}$, and a prior $\mathbf{D}$. Suppose that, under the game's hard distribution $\Delta_\theta$, the membership inference test $T$ has recall at least $q(n)$ and false-positive rate $O(\delta^2/n^2)$ with confidence $\delta$ (Definition~\ref{def:detection-rate}), and that $\delta = o(q(n))$. Then there is a reconstruction algorithm (Algorithm~\ref{alg:ra1}) that, with probability $1 - O(\delta)$, returns $S' \subseteq S$ with \(\E[|S'|] = \Omega(n \cdot q(n)).\)
\end{theorem}

\begin{algorithm}[t]
\caption{Basic Reconstruction}
\label{alg:ra1}
\small
\begin{algorithmic}[1]
\State \textbf{Input:} $\thetahat$, $\theta$, candidate set $\Sigma = \{\sigma_1, \ldots, \sigma_N\}$
\State \textbf{Output:} Reconstructed set $S'$
\State $S' \gets \emptyset$
\For{$i = 1, \ldots, N$}
    \If{$T(\thetahat, \theta, \sigma_i) = 1$}
        \State $S' \gets S' \cup \{\sigma_i\}$
    \EndIf
\EndFor
\State \textbf{return} $S'$
\end{algorithmic}
\end{algorithm}

\subsection{Proof of Theorem~\ref{thm:reduction}}

\paragraph{Proof overview.}
The key idea is to discretize the domain into a finite candidate set $\Sigma$,
small enough to search exhaustively but large enough to avoid collisions among
training samples (Steps~1--2). The membership test is then applied to every candidate: its low
false-positive rate ensures that only true training points are flagged (Step~3), while its
recall guarantees that many are recovered (Steps~4--5).

\paragraph{Step 1: Distribution construction.}
Steps~2--3 of the game (Definition~\ref{def:game}) are played with $N := |\Sigma| = n^2/\delta$: the adversary samples $\Sigma = \{\sigma_1, \ldots, \sigma_N\} \sim \cD_\theta^{\otimes N}$ from the ambient family and defines the \emph{discrete uniform distribution} \(\Delta_\theta := \text{Uniform}(\Sigma)\), from which the training sample $S \sim \Delta_\theta^{\otimes n}$ is drawn; the conditions of Definition~\ref{def:detection-rate} are required under $\Delta_\theta$.

\paragraph{Step 2: Collision bound.}
When $N \ge n^2/\delta$, i.i.d.\ sampling from $\Delta_\theta$ rarely produces collisions, so with high probability each training datapoint corresponds to a distinct element of $\Sigma$. Specifically, $\Prob[\text{each } \sigma_i \text{ sampled at most once}] = \prod_{k=0}^{n-1} \frac{N-k}{N} \ge (1 - \frac{n}{N})^n \ge (1 - \frac{\delta}{n})^n \ge 1 - \delta$, where the last inequality is Bernoulli's inequality $(1 - x)^n \ge 1 - nx$ for $x \in [0, 1]$.
Therefore, with probability at least $1 - \delta$, all elements of $S$ are distinct, so the flagged training points counted by condition~\eqref{eq:true-positive} correspond to as many distinct recovered candidates.

\paragraph{Step 3: No false positives.}
Writing $T(\sigma)$ for $T(\thetahat, \theta, \sigma)$, the union bound over the $N$ candidates and condition~\eqref{eq:false-positive} give
$$\begin{aligned}
&\Prob[\exists \sigma \in \Sigma : T(\sigma) = 1 \land \sigma \notin S] \\
&\quad\le N \cdot \max_{\sigma \in \Sigma} \Prob[T(\sigma) = 1 \mid \sigma \notin S] \\
&\quad\le O\left(N \cdot \frac{\delta^2}{n^2}\right)
= O(\delta).
\end{aligned}$$
Thus, with probability $1 - O(\delta)$, the test has no false positives within the candidate set $\Sigma$, ensuring $S' \subseteq S$.

\paragraph{Step 4: Reconstruction algorithm.}
Algorithm~\ref{alg:ra1} implements the reconstruction by exhaustively testing all candidates in $\Sigma$.

\paragraph{Step 5: Expected reconstruction size.}
Let $E$ denote the intersection of the no-collision event (Step~2, probability $\ge 1 - \delta$) and the no-false-positive event (Step~3, probability $\ge 1 - O(\delta)$), so $\Prob[E] \ge 1 - O(\delta)$. Let $X = |\{x \in S : T(x) = 1\}|$ be the number of flagged training points; the recall condition~\eqref{eq:true-positive} gives $\E[X] \ge q(n) \cdot n$. On the event $E$ the flagged training points are distinct and are all returned, so $|S'| \ge X - n \cdot \indicator[E^c]$ pointwise (trivially on $E^c$). Taking expectations,
\[
\begin{aligned}
\E[|S'|] &\;\ge\; \E[X] - n \cdot \Prob[E^c] \;\ge\; q(n) \cdot n - O(\delta) \cdot n \\
&\;=\; (1 - o(1))\, q(n) \cdot n \;=\; \Omega(n \cdot q(n)),
\end{aligned}
\]
where the first equality uses the hypothesis $\delta = o(q(n))$. This completes the proof.

\section{Reconstruction on Exponential Domains}
\label{sec:exponential}

The reduction in Section~\ref{sec:reduction} tests a candidate set $\Sigma$ exhaustively, so a polynomial runtime needs a polynomial-size set, whereas natural distributions have exponentially large support ($2^d$ for $d$-dimensional binary data). In this section we overcome this barrier with distributions partitioned into polynomially many sets, each containing exponentially many elements; we call these sets ``categories'', as the members of each share a common property.

We specialize to SCO, with the convex problem instance (Section~\ref{sec:reduction}) fixed and explicit: $\Theta = \{\theta \in \R^d : \|\theta\| \le 1\}$, $\cZ = \{\pm 1/\sqrt{d}\}^d$, and the linear (hence convex) loss $f(\theta, z) = -\langle \theta, z \rangle$, i.e.\ mean estimation, the standard hard instance of the SCO lower-bound literature \citep{attias2024information}. The parameter $\theta$ of Section~\ref{sec:reduction} becomes the distribution mean $\mu$, $\cD_\theta$ becomes the product Bernoulli family $\cD_\mu$, and the learner outputs $\thetahat \in \Theta$; the role of the hard distribution $\Delta_\theta$ of Definition~\ref{def:game} is played by the block-uniform $\Delta_\mu$ of Section~\ref{sec:construction}, under which both guarantees of Definition~\ref{def:detection-rate} are established.

\begin{definition}[Block-uniform distribution]
\label{def:block-uniform}
A distribution $\Delta$ over a domain $\cZ$ is \emph{block-uniform} with $N$ categories if its support can be partitioned into $N$ disjoint \emph{categories} $C_1, \ldots, C_N$, and $\Delta$ samples by drawing $l \sim \text{Uniform}([N])$ and then $z \sim \text{Uniform}(C_l)$.
\end{definition}

In our construction, $N = \text{poly}(n)$ while $|C_l| = 2^{d^{1/4}} - 2 \ge 2^{d^{1/4} - 1}$ for every $l$, so the support is exponentially large but organized into polynomially many categories.

\begin{theorem}[Impossibility over Exponential Domains]
\label{thm:exponential}
Let $(\Theta, \cZ, f)$ be the linear SCO instance above, and let $\cD_\mu$ be the product distribution over $\cZ$ with $\Prob[z_i = +1/\sqrt{d}] = \tfrac{1}{2}(1 + \mu_i\sqrt{d})$ independently for each coordinate, where $\mu \sim \text{Uniform}([-12\eps/\sqrt{d},\; 12\eps/\sqrt{d}]^d)$ so that $\E[z_i] = \mu_i$.

There exists $\eps_0 = 1/(12\sqrt{2})$ such that for every $\eps \le \eps_0$ and $\delta \le \eps$ with $\delta^{-1} \in \text{poly}(n)$, every $n$ in the regime $n = \Theta(1/\eps^2)$, and every $d \ge d_0 = \Omega\!\big(\max\{n^2 \log(n + \delta^{-1}),\; \eps^4 n^4/\delta^2,\; n^8 \log^4(n/\delta)\}\big)$, the following holds: for every $\eps$-learner $\cA_n$ there exists a block-uniform distribution $\Delta_\mu$, drawn from the explicitly constructed family of Section~\ref{sec:construction}, whose support consists of $N \cdot (2^{d^{1/4}} - 2)$ vectors ($N = n^2/\delta$ categories of $2^{d^{1/4}} - 2$ vectors each), such that, for a training set $S \sim \Delta_\mu^{\otimes n}$, with probability $\ge 1 - O(\delta)$, Algorithm~\ref{alg:ra2} reconstructs $\Omega(n)$ points of $S$ in $\text{poly}(n,d,\delta^{-1})$ time.
\end{theorem}

Here $n = \Theta(1/\eps^2)$ abbreviates the window $\Omega(1/\eps^2) \le n \le 1/(13824\,\eps^2)$ (Appendix~\ref{app:technical} explains both ends). The quantifier order matters: the hard distribution is chosen \emph{per learner}, and some such dependence is necessary at the standard sample size (Appendix~\ref{app:barrier}), whereas Algorithm~\ref{alg:ra2} is a \emph{single} explicit algorithm working against every $\eps$-learner. Sections~\ref{sec:construction}--\ref{sec:algorithm} together constitute the proof.

\subsection{Proof Overview and Construction}
\label{sec:construction}

\paragraph{Proof overview.}
The exhaustive search of Theorem~\ref{thm:reduction} is infeasible over the
exponentially large domain $\cZ$. We overcome this by partitioning the support into
polynomially many \emph{categories}, each with a rigid internal structure: some
coordinates are pinned to match the distribution mean, while the rest are
block-constant. This structure makes each category decodable, so given the learned
model, we can recover the unique training point in a category without enumerating it.
The proof combines four ingredients: (1)~the correlation bound
(Lemma~\ref{lem:correlation}, proven for product samples and applied to $\Delta_\mu$
in Appendix~\ref{app:complete-proof}) guarantees that most training points leave a detectable
signal in $\thetahat$; (2)~the category tail bound (Lemma~\ref{lem:category-tail})
ensures that non-training points do not produce false positives;
(3)~sign consistency (Lemma~\ref{lem:sign-consistency}) guarantees that the decoder
recovers each flagged training point exactly; and (4)~the algorithm runs in polynomial
time by iterating over the polynomially many categories.

The construction has two layers, fixed by the following partition structure. Each category $l \in [N]$ is governed by a partition of the coordinates
\[
\Pi_l \;=\; \big(\pi_{l,0},\; \pi_{l,1},\; \ldots,\; \pi_{l,d^{1/4}}\big),
\]
consisting of the \emph{mean-alignment part} $\pi_{l,0}$, whose coordinates are \emph{pinned} to fixed values chosen (by the bit-fitting rule below) so that the mean of the resulting distribution matches $\mu$, hence the name, and the $d^{1/4}$ \emph{blocks} $\pi_{l,1}, \ldots, \pi_{l,d^{1/4}}$, on which every member of category $l$ is constant. The intuition: pinned coordinates make $\E[z] \approx \mu$; blocks make categories decodable.

\paragraph{Partition sampling.}
For each category $l \in [N]$ ($N = n^2/\delta$), construct $\Pi_l = (\pi_{l,0}, \pi_{l,1}, \ldots, \pi_{l,d^{1/4}})$: include each $i \in [d]$ in $\pi_{l,0}$ with probability $\eps/\sqrt{d}$ independently; assign remaining coordinates uniformly to $d^{1/4}$ bins (Algorithm~\ref{alg:partition-sampling} in Appendix~\ref{app:algorithms}).

\paragraph{Bit-fitting.}
For each $i \in [d]$, let $K_i = \{l : i \in \pi_{l,0}\}$. Choose pin values $(B_{i,l})_{l \in K_i} \in \{\pm 1\}^{|K_i|}$ to minimize $|\mu_i - \frac{1}{N}\sum_{l \in K_i} B_{i,l}/\sqrt{d}|$ (Algorithm~\ref{alg:distribution-sampling} in Appendix~\ref{app:algorithms}).

\paragraph{Category definition.}
Category $C_l$ consists of all $z \in \{\pm 1/\sqrt{d}\}^d$ satisfying: (i) $z_i = B_{i,l}/\sqrt{d}$ for $i \in \pi_{l,0}$ (pinned), and (ii) $z_i = z_{i'}$ for all $i, i' \in \pi_{l,k}$, $k \in [d^{1/4}]$ (block-constant), excluding the two vectors whose $d^{1/4}$ block signs all coincide (all blocks positive, or all negative), so $|C_l| = 2^{d^{1/4}} - 2$. To sample $x \sim \Delta_\mu$: draw $l \sim \text{Uniform}([N])$, then $x \sim \text{Uniform}(C_l)$. The categories are pairwise disjoint except with probability $e^{-\Omega(\eps\sqrt{d})}$ over the partitions (Lemma~\ref{lem:disjointness}); disjointness can also be enforced deterministically by trimming, leaving Algorithm~\ref{alg:ra2} unchanged (Remark~\ref{rem:trimming}).

\paragraph{Mean-matching.}
For coordinate $i$: if $i \notin \pi_{l,0}$, the block sign is symmetric (the excluded pair is sign-flip symmetric) so $\E[x_i \mid l] = 0$; if $i \in \pi_{l,0}$, then $x_i = B_{i,l}/\sqrt{d}$. Averaging over $l$: $\E[x_i] = \frac{1}{N}\sum_{l \in K_i} B_{i,l}/\sqrt{d} \approx \mu_i$ by bit-fitting.

\subsection{Soundness of the Membership Test under \texorpdfstring{$\Delta_\mu$}{Delta\_mu}}
\label{sec:soundness}

This subsection verifies the false-positive condition~\eqref{eq:false-positive} of Definition~\ref{def:detection-rate} for the constructed $\Delta_\mu$, where the technical difficulty of this section is concentrated: $\Delta_\mu$ is \emph{not} a product distribution (coordinates within a block are perfectly correlated), so the product-measure sub-Gaussian soundness argument of \citet{attias2024information} does not apply. Lemma~\ref{lem:category-tail} and Corollary~\ref{cor:soundness} restore it; without them, the argument of Section~\ref{sec:reduction} could not be carried over to $\Delta_\mu$.

\begin{lemma}[Category tail bound]
\label{lem:category-tail}
Fix category $l$, draw $z \sim \text{Uniform}(C_l)$, and let $M_l = \max_k |\pi_{l,k}|$. For fixed $y \in \R^d$,
\[
\Prob\!\left[\langle y, z - \E[z \mid l] \rangle \ge t \right] \le 2\exp\!\left(-\frac{t^2 d}{2 M_l \|y\|^2}\right).
\]
\end{lemma}

\begin{proof}[Proof sketch]
Adding back the two excluded vectors makes the block signs independent Rademachers, so Hoeffding's inequality applies (Cauchy--Schwarz within each block yields the factor $M_l$); removing this sign-flip symmetric pair leaves $\E[z \mid l]$ unchanged and at most doubles the tail (full proof in Appendix~\ref{app:soundness-transfer}).
\end{proof}

\begin{corollary}[Soundness under $\Delta_\mu$]
\label{cor:soundness}
Let $A = \text{diag}(g(\eps,\mu))$ be the reweighting matrix of \citet[Lemma~6.3]{attias2024information} (restated as Definition~\ref{def:reweighting}). Setting $y = A\thetahat$ with $\|A\|_2 \le 288\eps^2$ (Lemma~\ref{lem:reweighting-norm}) and $\|\thetahat\| \le 1$, assuming $n = \Theta(1/\eps^2)$ as in Theorem~\ref{thm:exponential} (in particular $n \le 1/(13824\,\eps^2)$) and $d = \Omega(M \cdot n^2 \log(n/\delta))$ where $M = \max_l M_l$, this ensures \(\Prob\!\left[\exists i \in [n]: \langle A\thetahat, Z_i - \mu \rangle \ge \eps/(2n)\right] \le O(\delta^2/n^2)\),
for fresh $Z_1, \ldots, Z_n \sim \Delta_\mu$ (using the center-shift bound, Lemma~\ref{lem:center-shift}); the slack below the test's threshold $\eps/n$ is consumed in Appendix~\ref{app:sampling}.
\end{corollary}

Remark~\ref{rem:dimension} traces the three terms of $d_0$ to mean alignment, sign consistency, and soundness, and resolves the self-referential requirement on $d$ in Corollary~\ref{cor:soundness}, where the maximum block size $M$ itself depends on $d$.

\begin{algorithm}[t]
\caption{Partition-Based Reconstruction (with subroutines Algorithms~\ref{alg:partition-sampling}--\ref{alg:distribution-sampling}, App.~\ref{app:algorithms})}
\label{alg:ra2}
\small
\begin{algorithmic}[1]
\State \textbf{Input:} $\thetahat$; $\mu$; dimension $d$; $g(\eps,\mu)$ (Def.~\ref{def:reweighting}); partitions $(\Pi_l)_{l=1}^N$ and pin values $(B_{i,l})$, built by the adversary via Algorithms~\ref{alg:partition-sampling}--\ref{alg:distribution-sampling} (App.~\ref{app:algorithms})
\State \textbf{Output:} Reconstructed set $S'$
\State $S' \gets \emptyset$
\For{$l = 1, \ldots, N$}
    \State $\forall i \in \pi_{l,0}$: $z'_i \gets B_{i,l}/\sqrt{d}$
    \For{$k = 1, \ldots, d^{1/4}$}
        \State $\forall i \in \pi_{l,k}$:
        \State \quad $z'_i \gets \text{sgn}\!\left(\sum_{j \in \pi_{l,k}} g(\eps, \mu)_j\, \thetahat_j\right) / \sqrt{d}$
    \EndFor
    \If{$\langle g(\eps,\mu) \odot (z' - \mu),\; \thetahat \rangle \geq \eps/n$}
        \State $S' \gets S' \cup \{z'\}$
    \EndIf
\EndFor
\State \textbf{return} $S'$
\end{algorithmic}
\end{algorithm}

\subsection{The Reconstruction Algorithm}
\label{sec:algorithm}

For each category $l$, Algorithm~\ref{alg:ra2} constructs a candidate $z'$ by: (i) setting pinned coordinates to their pin values, and (ii) decoding block-constant coordinates via the sign of the reweighted block score $\sum_{j \in \pi_{l,k}} g(\eps,\mu)_j \thetahat_j$, where $g(\eps,\mu)_i$ is the coordinate-wise reweighting scalar defined in Definition~\ref{def:reweighting}, and $\odot$ denotes the Hadamard (entrywise) product. It then applies the membership test, here the thresholding rule $T(\thetahat, \mu, x) = \indicator[\langle g(\eps,\mu) \odot (x - \mu),\; \thetahat \rangle \ge \eps/n]$, which instantiates the abstract test $T: \Theta \times \Theta \times \cZ \to \{0, 1\}$ of Definition~\ref{def:mi-test}.

Appendix~\ref{sec:analysis} analyzes these ingredients, Appendix~\ref{app:proofs} gives the deferred proofs (the full sign-consistency derivation is Appendix~\ref{app:chebyshev-detailed}), and Appendix~\ref{app:complete-proof} assembles the complete proof.

\section{Related Work}
\label{sec:related}

\paragraph{Reconstruction attacks and traceability.}

Reconstruction from aggregate data is not a new topic: in an early result, \citet{dinur2003revealing} showed that answering too many linear queries on a database enables reconstruction of nearly the entire dataset -- a result that motivated the development of differential privacy; see \citet{dwork2017exposed} for a broad survey of reconstruction and re-identification attacks. In the learning setting, \citet{carlini2019secret} demonstrated unintended memorization in neural networks, and \citet{carlini2021extracting} extracted verbatim training data from large language models; \citet{nasr2023scalable} later scaled such extraction to production language models, while \citet{carlini2023diffusion} extracted training images from diffusion models.
On the theoretical side, \citet{balle2022reconstructing} analyzed reconstruction by informed adversaries with partial knowledge of the training set (e.g., all but one point), and \citet{haim2022reconstructing} developed reconstruction algorithms for neural network classifiers.
\citet{cohen2024reconstruction} proposed formal definitions and protection notions for reconstruction attacks, identifying when an attack should count as convincing evidence of vulnerability, and \citet{guo2022bounding} provided upper bounds on reconstruction success under DP, subsequently tightened by \citet{hayes2023bounding} for DP-SGD.

Closest to our setting, \citet{voitovych2025traceability} studied traceability in $\ell_p$ SCO, showing that $\eps$-learners are vulnerable to traceability attacks. Traceability -- identifying which individuals contributed to a dataset given aggregate statistics -- was formalized by \citet{dwork2015robust}, building on the genomic re-identification attacks of \citet{homer2008resolving}. The fingerprinting code technique underlying many traceability lower bounds originated with \citet{tardos2008optimal} and was adapted to differential privacy by \citet{bun2014fingerprinting} and \citet{steinke2017tight}. Our work goes beyond traceability and membership inference to establish full reconstruction: recovering the actual training datapoints themselves.

\paragraph{Membership inference.}

\citet{shokri2017membership} introduced membership inference attacks against machine learning models, using shadow models to distinguish training set members from non-members. \citet{yeom2018privacy} connected membership inference to the generalization gap, and \citet{nasr2019comprehensive} extended these attacks to the white-box and federated settings. \citet{sablayrolles2019white} derived Bayes-optimal membership inference strategies, and \citet{carlini2022membership} developed a principled likelihood-ratio framework characterizing when and why attacks succeed. \citet{choquette2021label} further showed that membership inference remains effective even when the adversary observes only hard label predictions. Our reduction (Theorem~\ref{thm:reduction}) converts any membership test that is sufficiently accurate under the hard distribution of Definition~\ref{def:game}, however constructed, into a reconstruction attack.

\paragraph{Memorization and generalization.}

Classical stability-based analyses by \citet{bousquet2002stability} established generalization guarantees for algorithms whose output changes little under single-sample perturbations, a notion closely tied to the memorization--leakage trade-off at the heart of our work. In SCO, \citet{shalev2010learnability} showed that learnability can hold even in the absence of uniform convergence, highlighting the distinctive role of algorithmic stability. \citet{zhang2017understanding} demonstrated that deep networks can fit random labels, challenging capacity-based explanations of generalization, and \citet{arpit2017closer} gave a complementary empirical characterization of memorization dynamics during training. \citet{feldman2020does} proved that memorization of individual examples is \emph{necessary} for near-optimal generalization on long-tailed distributions, and \citet{feldman2020what} validated this empirically via influence estimation. \citet{brown2021when} characterized when memorization of ``irrelevant'' training data is unavoidable for high-accuracy learning. \citet{carlini2023quantifying} measured how memorization scales with model size across language models.

\paragraph{Differential privacy and private SCO.}

Differential privacy \citep{dwork2006calibrating, dwork2014algorithmic} provides a rigorous framework for limiting information leakage from computations on private data. \citet{kasiviswanathan2011what} characterized what can be learned under differential privacy in the PAC model, and \citet{abadi2016deep} introduced DP-SGD, the standard practical tool for training neural networks under differential privacy. In the SCO setting, \citet{chaudhuri2011differentially} initiated the study of differentially private empirical risk minimization, and \citet{bassily2014private} established tight dimension-dependent excess-risk bounds for private convex ERM. \citet{jain2014near} showed that near dimension-independent rates are attainable under additional structural assumptions. \citet{bassily2019private} achieved optimal rates for private SCO, showing that the optimal excess population loss is the larger of the non-private SCO rate and the private-ERM rate, and \citet{feldman2020private} gave the first linear-time optimal algorithm. \citet{asi2021private} extended these results to $\ell_1$ geometry. Our results complement these bounds: every $\eps$-learner trained at the non-private sample size $n = \Theta(1/\eps^2)$ (with $d \ge d_0$) is vulnerable to reconstruction (Theorem~\ref{thm:exponential}).

\paragraph{Information complexity in SCO.}

SCO was studied by \citet{shalev2009stochastic}, who established its learnability structure and the failure of uniform convergence and ERM in the general case, while \citet{agarwal2012information} established information-theoretic oracle-complexity lower bounds. \citet{feldman2016generalization} showed that ERM requires dimension-dependent sample complexity, and \citet{livni2024sample} recently analyzed the sample complexity of gradient descent itself in SCO. Stability-based analyses of generalization -- pioneered by \citet{hardt2016train} for SGD -- provide a complementary perspective, with \citet{bassily2016algorithmic} establishing algorithmic stability as a unifying lens on adaptive data analysis.

Information-theoretic generalization bounds using mutual information were developed by \citet{russo2020much} and \citet{xu2017information}, and refined through the conditional mutual information framework of \citet{steinke2020reasoning}. \citet{haghifam2023limitations} identified limitations of these bounds for gradient descent in SCO, and \citet{livni2024information} proved information-theoretic lower bounds on the tightness of such bounds.

Most relevant to our work, \citet{attias2024information} combined these threads, proving that $\eps$-learners in SCO must memorize their training data in an information-theoretic sense, with the learned model necessarily revealing large conditional mutual information about individual training points. We build directly on this result and on their correlation bound (their Lemma~6.3): we establish the membership test's recall and soundness under a non-product block-uniform distribution, and show that such memorization enables not just membership inference but full reconstruction, even over exponentially large supports.

\section{Conclusion}
\label{sec:conclusion}

We established a general theoretical framework for reconstruction attacks on learning algorithms, with three main contributions. First, we presented a learner-agnostic reduction (Theorem~\ref{thm:reduction}) showing that any membership inference test with low false-positive rate and non-negligible recall under a hard distribution of polynomial support enables efficient reconstruction of training points drawn from it. Second, we developed a block-uniform distribution construction (Theorem~\ref{thm:exponential}) that enables polynomial-time reconstruction over exponentially large domains, using partition structure to reduce the search space from $2^d$ to $\text{poly}(n, d, \delta^{-1})$. Third, importing the correlation bound of \citet[Lemma~6.3]{attias2024information} and establishing our membership test's recall and soundness under the constructed block-uniform distribution itself, we showed that any $\eps$-learner with distribution-free guarantees is inherently vulnerable to reconstruction of $\Omega(n)$ training datapoints.

These results complete a chain of impossibility: \citet{attias2024information} proved that achieving low excess risk ($\eps$-learning) requires memorization of training data in an information-theoretic sense, and that this memorization enables membership inference. Our work shows that membership inference, in turn, enables reconstruction. Together, these results establish that \emph{in SCO, no learner can achieve low excess risk on all distributions while preventing reconstruction on all of them}; the obstacle lies in the learning problem, not in a particular algorithm.

These are worst-case statements: they do not say that a particular model trained on particular data is vulnerable, but that every $\eps$-learner with distribution-free guarantees is vulnerable on some distribution in the family.

\paragraph{Limitations.} The adversary chooses the hard distribution per learner and is given its parameter; the former is necessary at the standard sample size (Appendix~\ref{app:barrier}), the latter relaxes to sampling access (Appendix~\ref{app:sampling}). Our analysis assumes the SCO setting with convex losses and the product Bernoulli family as the ambient family (the hard distribution itself is block-uniform); extensions to non-convex settings or more general distribution families remain open. The dimension requirement $d \ge \Omega(\max\{n^2 \log(n + \delta^{-1}),\; \eps^4 n^4/\delta^2,\; n^8 \log^4(n/\delta)\})$, whose third, soundness-driven term dominates for $\delta^{-1} = O(n^3)$ (Remark~\ref{rem:dimension}), is large in practice and grows with the sample size, and Theorem~\ref{thm:exponential} operates in the regime $n = \Theta(1/\eps^2)$. Our results are asymptotic, and the constants hidden in $\Omega(\cdot)$ and $O(\cdot)$ notation matter for practical attacks.

\paragraph{Future directions.} Several directions merit further investigation: (1) the conditions under which reconstructions are preventable; (2) the effects of adaptive defenses and their fundamental limitations; and (3) technical improvements to our framework, including tighter dimension bounds that reduce the polynomial dependence on $n$, formal characterization of when natural data distributions exhibit block-uniform structure, and precise lower bounds on the privacy--accuracy trade-off. The extension we consider most promising, to arbitrary exponential families via the generalized fingerprinting lemma of \citet{kamath2022new}, is stated as an open problem in Appendix~\ref{app:exp-families}; Appendix~\ref{app:discussion} also treats continuous distributions (Appendix~\ref{app:continuous}) and the category axioms (Appendix~\ref{app:axioms}).

\subsection*{AI use statement}

In this work, we used generative AI tools for assisting in the writing of proofs and for assisting with translation.
We have not used generative AI tools for helping develop theoretical models or conceptual frameworks, formulating mathematical claims, providing critical ingredients for proving mathematical claims, or proposing or refining hypotheses; generating synthetic data sets, designing or providing feedback on research methodology or experiments, implementing methods, cleaning and reformatting datasets, supporting qualitative and thematic data analysis, and interpreting results are not applicable to this work.
Additionally, we used generative AI tools for searching for relevant literature, editing phrases to improve readability, and formatting references.
We have reviewed all AI-assisted work.
All AI-assisted proof text was checked line by line by the authors.
We take responsibility for the final content of this work, including text, claims, or artifacts produced with the aid of generative AI.

\acknowledgments{
We would like to thank Roi Livni for brilliant discussions about his membership inference work and how to expand from it. We additionally thank Yishay Mansour for his insightful advice for reconstruction schemes.
}

\bibliographystyle{plainnat}
\bibliography{references}

\begin{thebibliography}{59}
\providecommand{\natexlab}[1]{#1}
\providecommand{\url}[1]{\texttt{#1}}
\expandafter\ifx\csname urlstyle\endcsname\relax
  \providecommand{\doi}[1]{doi: #1}\else
  \providecommand{\doi}{doi: \begingroup \urlstyle{rm}\Url}\fi

\bibitem[Abadi et~al.(2016)Abadi, Chu, Goodfellow, McMahan, Mironov, Talwar,
  and Zhang]{abadi2016deep}
Mart{\'\i}n Abadi, Andy Chu, Ian Goodfellow, H~Brendan McMahan, Ilya Mironov,
  Kunal Talwar, and Li~Zhang.
\newblock Deep learning with differential privacy.
\newblock In \emph{ACM SIGSAC Conference on Computer and Communications
  Security (CCS)}, pages 308--318, 2016.

\bibitem[Agarwal et~al.(2012)Agarwal, Bartlett, Ravikumar, and
  Wainwright]{agarwal2012information}
Alekh Agarwal, Peter~L Bartlett, Pradeep Ravikumar, and Martin~J Wainwright.
\newblock Information-theoretic lower bounds on the oracle complexity of
  stochastic convex optimization.
\newblock \emph{IEEE Transactions on Information Theory}, 58\penalty0
  (5):\penalty0 3235--3249, 2012.

\bibitem[Arpit et~al.(2017)Arpit, Jastrz{\k{e}}bski, Ballas, Krueger, Bengio,
  Kanwal, Maharaj, Fischer, Courville, Bengio, et~al.]{arpit2017closer}
Devansh Arpit, Stanis{\l}aw Jastrz{\k{e}}bski, Nicolas Ballas, David Krueger,
  Emmanuel Bengio, Maxinder~S Kanwal, Tegan Maharaj, Asja Fischer, Aaron
  Courville, Yoshua Bengio, et~al.
\newblock A closer look at memorization in deep networks.
\newblock In \emph{International Conference on Machine Learning (ICML)}, pages
  233--242. PMLR, 2017.

\bibitem[Asi et~al.(2021)Asi, Feldman, Koren, and Talwar]{asi2021private}
Hilal Asi, Vitaly Feldman, Tomer Koren, and Kunal Talwar.
\newblock Private stochastic convex optimization: Optimal rates in $\ell_1$
  geometry.
\newblock In \emph{International Conference on Machine Learning (ICML)}, pages
  393--403. PMLR, 2021.

\bibitem[Ateniese et~al.(2015)Ateniese, Felici, Mancini, Spognardi, Villani,
  and Vitali]{ateniese2015hacking}
Giuseppe Ateniese, Giovanni Felici, Luigi~V Mancini, Angelo Spognardi, Antonio
  Villani, and Domenico Vitali.
\newblock Hacking smart machines with smarter ones: How to extract meaningful
  data from machine learning classifiers.
\newblock \emph{International Journal of Security and Networks}, 10\penalty0
  (3):\penalty0 137--150, 2015.

\bibitem[Attias et~al.(2024)Attias, Dziugaite, Haghifam, Livni, and
  Roy]{attias2024information}
Idan Attias, Gintare~Karolina Dziugaite, Mahdi Haghifam, Roi Livni, and
  Daniel~M. Roy.
\newblock Information complexity of stochastic convex optimization:
  Applications to generalization and memorization.
\newblock arXiv:2402.09327v2, 2024.
\newblock Conference version in ICML 2024.

\bibitem[Balle et~al.(2022)Balle, Cherubin, and Hayes]{balle2022reconstructing}
Borja Balle, Giovanni Cherubin, and Jamie Hayes.
\newblock Reconstructing training data with informed adversaries.
\newblock In \emph{IEEE Symposium on Security and Privacy (SP)}, pages
  1138--1156. IEEE, 2022.

\bibitem[Bassily et~al.(2014)Bassily, Smith, and Thakurta]{bassily2014private}
Raef Bassily, Adam Smith, and Abhradeep Thakurta.
\newblock Private empirical risk minimization: Efficient algorithms and tight
  error bounds.
\newblock In \emph{IEEE Symposium on Foundations of Computer Science (FOCS)},
  pages 464--473. IEEE, 2014.

\bibitem[Bassily et~al.(2016)Bassily, Nissim, Smith, Steinke, Stemmer, and
  Ullman]{bassily2016algorithmic}
Raef Bassily, Kobbi Nissim, Adam Smith, Thomas Steinke, Uri Stemmer, and
  Jonathan Ullman.
\newblock Algorithmic stability for adaptive data analysis.
\newblock In \emph{ACM SIGACT Symposium on Theory of Computing (STOC)}, pages
  1046--1059, 2016.

\bibitem[Bassily et~al.(2019)Bassily, Feldman, Talwar, and
  Guha~Thakurta]{bassily2019private}
Raef Bassily, Vitaly Feldman, Kunal Talwar, and Abhradeep Guha~Thakurta.
\newblock Private stochastic convex optimization with optimal rates.
\newblock In \emph{Advances in Neural Information Processing Systems
  (NeurIPS)}, volume~32, 2019.

\bibitem[Bousquet and Elisseeff(2002)]{bousquet2002stability}
Olivier Bousquet and Andr{\'e} Elisseeff.
\newblock Stability and generalization.
\newblock \emph{Journal of Machine Learning Research}, 2:\penalty0 499--526,
  2002.

\bibitem[Brown et~al.(2021)Brown, Bun, Feldman, Smith, and
  Talwar]{brown2021when}
Gavin Brown, Mark Bun, Vitaly Feldman, Adam Smith, and Kunal Talwar.
\newblock When is memorization of irrelevant training data necessary for
  high-accuracy learning?
\newblock In \emph{ACM SIGACT Symposium on Theory of Computing (STOC)}, pages
  123--132, 2021.

\bibitem[Brown et~al.(2022)Brown, Lee, Mireshghallah, Shokri, and
  Tram{\`e}r]{brown2022language}
Hannah Brown, Katherine Lee, Fatemehsadat Mireshghallah, Reza Shokri, and
  Florian Tram{\`e}r.
\newblock What does it mean for a language model to preserve privacy?
\newblock In \emph{2022 ACM Conference on Fairness, Accountability, and
  Transparency (FAccT)}, pages 2280--2292. ACM, 2022.

\bibitem[Bun et~al.(2014)Bun, Ullman, and Vadhan]{bun2014fingerprinting}
Mark Bun, Jonathan Ullman, and Salil Vadhan.
\newblock Fingerprinting codes and the price of approximate differential
  privacy.
\newblock In \emph{ACM SIGACT Symposium on Theory of Computing (STOC)}, pages
  1--10, 2014.

\bibitem[Carlini et~al.(2019)Carlini, Liu, Erlingsson, Kos, and
  Song]{carlini2019secret}
Nicholas Carlini, Chang Liu, {\'U}lfar Erlingsson, Jernej Kos, and Dawn Song.
\newblock The secret sharer: Evaluating and testing unintended memorization in
  neural networks.
\newblock In \emph{USENIX Security Symposium}, pages 267--284, 2019.

\bibitem[Carlini et~al.(2021)Carlini, Tram{\`e}r, Wallace, Jagielski,
  Herbert-Voss, Lee, Roberts, Brown, Song, Erlingsson,
  et~al.]{carlini2021extracting}
Nicholas Carlini, Florian Tram{\`e}r, Eric Wallace, Matthew Jagielski, Ariel
  Herbert-Voss, Katherine Lee, Adam Roberts, Tom Brown, Dawn Song, {\'U}lfar
  Erlingsson, et~al.
\newblock Extracting training data from large language models.
\newblock In \emph{USENIX Security Symposium}, pages 2633--2650, 2021.

\bibitem[Carlini et~al.(2022)Carlini, Chien, Nasr, Song, Terzis, and
  Tram{\`e}r]{carlini2022membership}
Nicholas Carlini, Steve Chien, Milad Nasr, Shuang Song, Andreas Terzis, and
  Florian Tram{\`e}r.
\newblock Membership inference attacks from first principles.
\newblock In \emph{2022 IEEE Symposium on Security and Privacy (SP)}, pages
  1897--1914. IEEE, 2022.

\bibitem[Carlini et~al.(2023{\natexlab{a}})Carlini, Hayes, Nasr, Jagielski,
  Sehwag, Tram{\`e}r, Balle, Ippolito, and Wallace]{carlini2023diffusion}
Nicholas Carlini, Jamie Hayes, Milad Nasr, Matthew Jagielski, Vikash Sehwag,
  Florian Tram{\`e}r, Borja Balle, Daphne Ippolito, and Eric Wallace.
\newblock Extracting training data from diffusion models.
\newblock In \emph{USENIX Security Symposium}, pages 5253--5270,
  2023{\natexlab{a}}.

\bibitem[Carlini et~al.(2023{\natexlab{b}})Carlini, Ippolito, Jagielski, Lee,
  Tram{\`e}r, and Zhang]{carlini2023quantifying}
Nicholas Carlini, Daphne Ippolito, Matthew Jagielski, Katherine Lee, Florian
  Tram{\`e}r, and Chiyuan Zhang.
\newblock Quantifying memorization across neural language models.
\newblock In \emph{International Conference on Learning Representations
  (ICLR)}, 2023{\natexlab{b}}.

\bibitem[Chaudhuri et~al.(2011)Chaudhuri, Monteleoni, and
  Sarwate]{chaudhuri2011differentially}
Kamalika Chaudhuri, Claire Monteleoni, and Anand~D Sarwate.
\newblock Differentially private empirical risk minimization.
\newblock \emph{Journal of Machine Learning Research}, 12:\penalty0 1069--1109,
  2011.

\bibitem[Choquette-Choo et~al.(2021)Choquette-Choo, Tram{\`e}r, Carlini, and
  Papernot]{choquette2021label}
Christopher~A Choquette-Choo, Florian Tram{\`e}r, Nicholas Carlini, and Nicolas
  Papernot.
\newblock Label-only membership inference attacks.
\newblock In \emph{International Conference on Machine Learning (ICML)}, pages
  1964--1974. PMLR, 2021.

\bibitem[Cohen et~al.(2024)Cohen, Kaplan, Mansour, Moran, Nissim, Stemmer, and
  Tsfadia]{cohen2024reconstruction}
Edith Cohen, Haim Kaplan, Yishay Mansour, Shay Moran, Kobbi Nissim, Uri
  Stemmer, and Eliad Tsfadia.
\newblock Data reconstruction: When you see it and when you don't.
\newblock \emph{arXiv preprint arXiv:2405.15753}, 2024.

\bibitem[Dinur and Nissim(2003)]{dinur2003revealing}
Irit Dinur and Kobbi Nissim.
\newblock Revealing information while preserving privacy.
\newblock In \emph{ACM SIGMOD-SIGACT-SIGART Symposium on Principles of Database
  Systems (PODS)}, pages 202--210, 2003.

\bibitem[Dwork and Roth(2014)]{dwork2014algorithmic}
Cynthia Dwork and Aaron Roth.
\newblock The algorithmic foundations of differential privacy.
\newblock \emph{Foundations and Trends in Theoretical Computer Science},
  9\penalty0 (3--4):\penalty0 211--407, 2014.

\bibitem[Dwork et~al.(2006)Dwork, McSherry, Nissim, and
  Smith]{dwork2006calibrating}
Cynthia Dwork, Frank McSherry, Kobbi Nissim, and Adam Smith.
\newblock Calibrating noise to sensitivity in private data analysis.
\newblock In \emph{Theory of Cryptography Conference (TCC)}, pages 265--284.
  Springer, 2006.

\bibitem[Dwork et~al.(2015)Dwork, Smith, Steinke, Ullman, and
  Vadhan]{dwork2015robust}
Cynthia Dwork, Adam Smith, Thomas Steinke, Jonathan Ullman, and Salil Vadhan.
\newblock Robust traceability from trace amounts.
\newblock In \emph{IEEE Symposium on Foundations of Computer Science (FOCS)},
  pages 650--669. IEEE, 2015.

\bibitem[Dwork et~al.(2017)Dwork, Smith, Steinke, and Ullman]{dwork2017exposed}
Cynthia Dwork, Adam Smith, Thomas Steinke, and Jonathan Ullman.
\newblock Exposed! {A} survey of attacks on private data.
\newblock \emph{Annual Review of Statistics and Its Application}, 4:\penalty0
  61--84, 2017.

\bibitem[Feldman(2016)]{feldman2016generalization}
Vitaly Feldman.
\newblock Generalization of {ERM} in stochastic convex optimization: The
  dimension strikes back.
\newblock In \emph{Advances in Neural Information Processing Systems
  (NeurIPS)}, volume~29, 2016.

\bibitem[Feldman(2020)]{feldman2020does}
Vitaly Feldman.
\newblock Does learning require memorization? {A} short tale about a long tail.
\newblock In \emph{ACM SIGACT Symposium on Theory of Computing (STOC)}, pages
  954--959, 2020.

\bibitem[Feldman and Zhang(2020)]{feldman2020what}
Vitaly Feldman and Chiyuan Zhang.
\newblock What neural networks memorize and why: Discovering the long tail via
  influence estimation.
\newblock In \emph{Advances in Neural Information Processing Systems
  (NeurIPS)}, volume~33, pages 2881--2891, 2020.

\bibitem[Feldman et~al.(2020)Feldman, Koren, and Talwar]{feldman2020private}
Vitaly Feldman, Tomer Koren, and Kunal Talwar.
\newblock Private stochastic convex optimization: Optimal rates in linear time.
\newblock In \emph{ACM SIGACT Symposium on Theory of Computing (STOC)}, pages
  439--449, 2020.

\bibitem[Fredrikson et~al.(2015)Fredrikson, Jha, and
  Ristenpart]{fredrikson2015model}
Matt Fredrikson, Somesh Jha, and Thomas Ristenpart.
\newblock Model inversion attacks that exploit confidence information and basic
  countermeasures.
\newblock In \emph{ACM SIGSAC Conference on Computer and Communications
  Security (CCS)}, pages 1322--1333, 2015.

\bibitem[Fredrikson et~al.(2014)Fredrikson, Lantz, Jha, Lin, Page, and
  Ristenpart]{fredrikson2014privacy}
Matthew Fredrikson, Eric Lantz, Somesh Jha, Simon Lin, David Page, and Thomas
  Ristenpart.
\newblock Privacy in pharmacogenetics: An end-to-end case study of personalized
  warfarin dosing.
\newblock In \emph{USENIX Security Symposium}, pages 17--32, 2014.

\bibitem[Ganju et~al.(2018)Ganju, Wang, Yang, Gunter, and
  Borisov]{ganju2018property}
Karan Ganju, Qi~Wang, Wei Yang, Carl~A Gunter, and Nikita Borisov.
\newblock Property inference attacks on fully connected neural networks using
  permutation invariant representations.
\newblock In \emph{ACM SIGSAC Conference on Computer and Communications
  Security (CCS)}, pages 619--633, 2018.

\bibitem[Guo et~al.(2022)Guo, Karrer, Chaudhuri, and van~der
  Maaten]{guo2022bounding}
Chuan Guo, Brian Karrer, Kamalika Chaudhuri, and Laurens van~der Maaten.
\newblock Bounding training data reconstruction in private (deep) learning.
\newblock In \emph{International Conference on Machine Learning (ICML)}, pages
  8056--8071. PMLR, 2022.

\bibitem[Haghifam et~al.(2023)Haghifam, Rodr{\'\i}guez-G{\'a}lvez, Thobaben,
  Skoglund, Roy, and Dziugaite]{haghifam2023limitations}
Mahdi Haghifam, Borja Rodr{\'\i}guez-G{\'a}lvez, Ragnar Thobaben, Mikael
  Skoglund, Daniel~M Roy, and Gintare~Karolina Dziugaite.
\newblock Limitations of information-theoretic generalization bounds for
  gradient descent methods in stochastic convex optimization.
\newblock In \emph{International Conference on Algorithmic Learning Theory
  (ALT)}, pages 663--706. PMLR, 2023.

\bibitem[Haim et~al.(2022)Haim, Vardi, Yehudai, Shamir, and
  Irani]{haim2022reconstructing}
Niv Haim, Gal Vardi, Gilad Yehudai, Ohad Shamir, and Michal Irani.
\newblock Reconstructing training data from trained neural networks.
\newblock In \emph{Advances in Neural Information Processing Systems
  (NeurIPS)}, volume~35, pages 22911--22924, 2022.

\bibitem[Hardt et~al.(2016)Hardt, Recht, and Singer]{hardt2016train}
Moritz Hardt, Benjamin Recht, and Yoram Singer.
\newblock Train faster, generalize better: Stability of stochastic gradient
  descent.
\newblock In \emph{International Conference on Machine Learning (ICML)}, pages
  1225--1234. PMLR, 2016.

\bibitem[Hayes et~al.(2023)Hayes, Mahloujifar, and Balle]{hayes2023bounding}
Jamie Hayes, Saeed Mahloujifar, and Borja Balle.
\newblock Bounding training data reconstruction in {DP-SGD}.
\newblock In \emph{Advances in Neural Information Processing Systems
  (NeurIPS)}, 2023.

\bibitem[Homer et~al.(2008)Homer, Szelinger, Redman, Duggan, Tembe, Muehling,
  Pearson, Stephan, Nelson, and Craig]{homer2008resolving}
Nils Homer, Szabolcs Szelinger, Margot Redman, David Duggan, Waibhav Tembe,
  Jill Muehling, John~V Pearson, Dietrich~A Stephan, Stanley~F Nelson, and
  David~W Craig.
\newblock Resolving individuals contributing trace amounts of {DNA} to highly
  complex mixtures using high-density {SNP} genotyping microarrays.
\newblock \emph{PLoS Genetics}, 4\penalty0 (8):\penalty0 e1000167, 2008.

\bibitem[Jain and Thakurta(2014)]{jain2014near}
Prateek Jain and Abhradeep Thakurta.
\newblock {(Near)} dimension independent risk bounds for differentially private
  learning.
\newblock In \emph{International Conference on Machine Learning (ICML)}, pages
  476--484. PMLR, 2014.

\bibitem[Kamath et~al.(2022)Kamath, Mouzakis, and Singhal]{kamath2022new}
Gautam Kamath, Argyris Mouzakis, and Vikrant Singhal.
\newblock New lower bounds for private estimation and a generalized
  fingerprinting lemma.
\newblock In \emph{Advances in Neural Information Processing Systems
  (NeurIPS)}, 2022.

\bibitem[Kasiviswanathan et~al.(2011)Kasiviswanathan, Lee, Nissim,
  Raskhodnikova, and Smith]{kasiviswanathan2011what}
Shiva~Prasad Kasiviswanathan, Homin~K Lee, Kobbi Nissim, Sofya Raskhodnikova,
  and Adam Smith.
\newblock What can we learn privately?
\newblock \emph{SIAM Journal on Computing}, 40\penalty0 (3):\penalty0 793--826,
  2011.

\bibitem[Livni(2023)]{livni2024information}
Roi Livni.
\newblock Information theoretic lower bounds for information theoretic upper
  bounds.
\newblock \emph{Advances in Neural Information Processing Systems (NeurIPS)},
  36, 2023.

\bibitem[Livni(2024)]{livni2024sample}
Roi Livni.
\newblock The sample complexity of gradient descent in stochastic convex
  optimization.
\newblock In \emph{Advances in Neural Information Processing Systems
  (NeurIPS)}, volume~37, 2024.

\bibitem[Nasr et~al.(2019)Nasr, Shokri, and Houmansadr]{nasr2019comprehensive}
Milad Nasr, Reza Shokri, and Amir Houmansadr.
\newblock Comprehensive privacy analysis of deep learning: Passive and active
  white-box inference attacks against centralized and federated learning.
\newblock In \emph{IEEE Symposium on Security and Privacy (SP)}, pages
  739--753. IEEE, 2019.

\bibitem[Nasr et~al.(2023)Nasr, Carlini, Hayase, Jagielski, Cooper, Ippolito,
  Choquette-Choo, Wallace, Tram{\`e}r, and Lee]{nasr2023scalable}
Milad Nasr, Nicholas Carlini, Jonathan Hayase, Matthew Jagielski, A~Feder
  Cooper, Daphne Ippolito, Christopher~A Choquette-Choo, Eric Wallace, Florian
  Tram{\`e}r, and Katherine Lee.
\newblock Scalable extraction of training data from (production) language
  models.
\newblock In \emph{arXiv preprint arXiv:2311.17035}, 2023.

\bibitem[Russo and Zou(2020)]{russo2020much}
Daniel Russo and James Zou.
\newblock How much does your data exploration overfit? {C}ontrolling bias via
  information usage.
\newblock \emph{IEEE Transactions on Information Theory}, 66\penalty0
  (1):\penalty0 302--323, 2020.

\bibitem[Sablayrolles et~al.(2019)Sablayrolles, Douze, Schmid, Ollivier, and
  J{\'e}gou]{sablayrolles2019white}
Alexandre Sablayrolles, Matthijs Douze, Cordelia Schmid, Yann Ollivier, and
  Herv{\'e} J{\'e}gou.
\newblock White-box vs black-box: {B}ayes optimal strategies for membership
  inference.
\newblock In \emph{International Conference on Machine Learning (ICML)}, pages
  5558--5567. PMLR, 2019.

\bibitem[Shalev-Shwartz et~al.(2009)Shalev-Shwartz, Shamir, Srebro, and
  Sridharan]{shalev2009stochastic}
Shai Shalev-Shwartz, Ohad Shamir, Nathan Srebro, and Karthik Sridharan.
\newblock Stochastic convex optimization.
\newblock In \emph{Conference on Learning Theory (COLT)}, 2009.

\bibitem[Shalev-Shwartz et~al.(2010)Shalev-Shwartz, Shamir, Srebro, and
  Sridharan]{shalev2010learnability}
Shai Shalev-Shwartz, Ohad Shamir, Nathan Srebro, and Karthik Sridharan.
\newblock Learnability, stability and uniform convergence.
\newblock \emph{Journal of Machine Learning Research}, 11:\penalty0 2635--2670,
  2010.

\bibitem[Shokri et~al.(2017)Shokri, Stronati, Song, and
  Shmatikov]{shokri2017membership}
Reza Shokri, Marco Stronati, Congzheng Song, and Vitaly Shmatikov.
\newblock Membership inference attacks against machine learning models.
\newblock In \emph{IEEE Symposium on Security and Privacy (SP)}, pages 3--18.
  IEEE, 2017.

\bibitem[Steinke and Ullman(2017)]{steinke2017tight}
Thomas Steinke and Jonathan Ullman.
\newblock Tight lower bounds for differentially private selection.
\newblock In \emph{IEEE Symposium on Foundations of Computer Science (FOCS)},
  pages 552--563. IEEE, 2017.

\bibitem[Steinke and Zakynthinou(2020)]{steinke2020reasoning}
Thomas Steinke and Lydia Zakynthinou.
\newblock Reasoning about generalization via conditional mutual information.
\newblock In \emph{Conference on Learning Theory (COLT)}, pages 3437--3452.
  PMLR, 2020.

\bibitem[Tardos(2008)]{tardos2008optimal}
G{\'a}bor Tardos.
\newblock Optimal probabilistic fingerprint codes.
\newblock \emph{Journal of the ACM}, 55\penalty0 (2):\penalty0 1--24, 2008.

\bibitem[Voitovych et~al.(2025)Voitovych, Haghifam, Attias, Dziugaite, Livni,
  and Roy]{voitovych2025traceability}
Sasha Voitovych, Mahdi Haghifam, Idan Attias, Gintare~Karolina Dziugaite, Roi
  Livni, and Daniel~M. Roy.
\newblock On traceability in $\ell_p$ stochastic convex optimization.
\newblock \emph{arXiv preprint arXiv:2502.17384}, 2025.

\bibitem[Xu and Raginsky(2017)]{xu2017information}
Aolin Xu and Maxim Raginsky.
\newblock Information-theoretic analysis of generalization capability of
  learning algorithms.
\newblock In \emph{Advances in Neural Information Processing Systems
  (NeurIPS)}, volume~30, pages 2524--2533, 2017.

\bibitem[Yeom et~al.(2018)Yeom, Giacomelli, Fredrikson, and
  Jha]{yeom2018privacy}
Samuel Yeom, Irene Giacomelli, Matt Fredrikson, and Somesh Jha.
\newblock Privacy risk in machine learning: Analyzing the connection to
  overfitting.
\newblock In \emph{IEEE Computer Security Foundations Symposium (CSF)}, pages
  268--282. IEEE, 2018.

\bibitem[Zhang et~al.(2017)Zhang, Bengio, Hardt, Recht, and
  Vinyals]{zhang2017understanding}
Chiyuan Zhang, Samy Bengio, Moritz Hardt, Benjamin Recht, and Oriol Vinyals.
\newblock Understanding deep learning requires rethinking generalization.
\newblock In \emph{International Conference on Learning Representations
  (ICLR)}, 2017.

\end{thebibliography}

\clearpage
\appendix
\thispagestyle{empty}
\onecolumn
\iflatexml
\else
\aistatstitle{\papertitle: \\
Supplementary Materials}
\fi

\section{Additional Algorithms}
\label{app:algorithms}

The main text presents two reconstruction algorithms: Algorithm~\ref{alg:ra1} (Section~\ref{sec:reduction}) for polynomial-sized candidate sets, and Algorithm~\ref{alg:ra2} (Section~\ref{sec:exponential}) for exponential domains. Below are the partition and distribution sampling subroutines used by Algorithm~\ref{alg:ra2}.

\subsection{Partition Sampling}

\begin{algorithm}[H]
\caption{Partition Sampling}
\label{alg:partition-sampling}
\begin{algorithmic}[1]
\State \textbf{Input:} Dimension $d$, number of categories $N$, parameter $\eps$
\State \textbf{Output:} Partitions $\Pi_1, \ldots, \Pi_N$
\For{$l = 1, \ldots, N$}
    \State $\pi_{l,0} \gets \emptyset$
    \For{$i = 1, \ldots, d$}
        \If{$\text{Bern}(\eps/\sqrt{d}) = 1$}
            \State $\pi_{l,0} \gets \pi_{l,0} \cup \{i\}$
        \EndIf
    \EndFor
    \State Initialize $\pi_{l,1}, \ldots, \pi_{l,d^{1/4}} \gets \emptyset$
    \For{$i \in [d] \setminus \pi_{l,0}$}
        \State $\text{bin} \sim \text{Uniform}(\{1, \ldots, d^{1/4}\})$;\ \ $\pi_{l,\text{bin}} \gets \pi_{l,\text{bin}} \cup \{i\}$
    \EndFor
    \State $\Pi_l \gets (\pi_{l,0}, \pi_{l,1}, \ldots, \pi_{l,d^{1/4}})$
\EndFor
\State \textbf{return} $\{\Pi_1, \ldots, \Pi_N\}$
\end{algorithmic}
\end{algorithm}

\subsection{Distribution Sampling}

\begin{algorithm}[H]
\caption{Distribution Sampling (Bit-Fitting + Category Definition)}
\label{alg:distribution-sampling}
\begin{algorithmic}[1]
\State \textbf{Input:} Mean $\mu \in \R^d$, partitions $\{\Pi_l\}_{l=1}^N$
\State \textbf{Output:} Pin values $(B_{i,l})$, category sets $\{C_1, \ldots, C_N\}$
\For{$i = 1, \ldots, d$}
    \State $K_i \gets \{l \in [N] : i \in \pi_{l,0}\}$
    \State Choose $(B_{i,l})_{l \in K_i} \in \{\pm 1\}^{|K_i|}$ minimizing $|\mu_i - \frac{1}{N} \sum_{l \in K_i} B_{i,l}/\sqrt{d}|$
\EndFor
\For{$l = 1, \ldots, N$}
    \State $C_l \gets \{z \in \{\pm 1/\sqrt{d}\}^d : (\forall i \in \pi_{l,0},\; z_i = B_{i,l}/\sqrt{d})$
    \Statex \hspace{4em} $\land\; (\forall k \in [d^{1/4}],\; \forall i, i' \in \pi_{l,k},\; z_i = z_{i'})$
    \Statex \hspace{4em} $\land\; \text{not all block signs of } z \text{ coincide}\}$
\EndFor
\State \textbf{return} $(B_{i,l})$, $\{C_1, \ldots, C_N\}$
\end{algorithmic}
\end{algorithm}

\textbf{Complexity of Algorithm~\ref{alg:ra2}:} $O(N \cdot d) = O(n^2 d / \delta)$, polynomial when $d = \text{poly}(n, 1/\delta)$.

\section{Technical Analysis}
\label{sec:analysis}

This section establishes the key technical ingredients for Theorem~\ref{thm:exponential}: the correlation bound, the reweighting vector, and sign consistency of block scores. Full proofs are deferred to Appendix~\ref{app:proofs}.

\subsection{Reweighting Vector and Correlation}

\begin{definition}[Reweighting vector]
\label{def:reweighting}
For $\eps > 0$ and $\mu \in [-12\eps/\sqrt{d}, 12\eps/\sqrt{d}]^d$, define
\[
g(\eps, \mu)_i = \frac{144\eps^2 - \mu_i^2 d}{1 - \mu_i^2 d}, \qquad v(z)_i = g(\eps, \mu)_i (z_i - \mu_i).
\]
\end{definition}

\begin{lemma}[Correlation bound {\citep{attias2024information}}]
\label{lem:correlation}
Under the uniform prior $U = \text{Uniform}([-12\eps/\sqrt{d},\allowbreak 12\eps/\sqrt{d}]^d)$, for any $\eps$-learner $\cA_n$, \(\E_{\mu \sim U, S_n \sim \cD_\mu^{\otimes n}}\!\left[\sum_{j=1}^n \langle v(z_j), \thetahat \rangle\right] \ge 6\eps - 4\delta\).
In particular, with probability $\ge1 - \delta$, at least $\Omega(n)$ training datapoints satisfy $\langle v(z_j), \thetahat \rangle \ge \eps/n$.
\end{lemma}

\begin{lemma}[Reweighting norm bound]
\label{lem:reweighting-norm}
For $\eps \le \eps_0 = 1/(12\sqrt{2})$ and any $J \subseteq [d]$: $|g(\eps,\mu)_i| \le 288\eps^2$ and $\sum_{i \in J} g(\eps,\mu)_i^2 \le 288^2 \eps^4 |J|$.
\end{lemma}

\begin{proof}
See Appendix~\ref{app:proofs}.
\end{proof}

\subsection{Sign Consistency}

The reconstruction algorithm sets block $k$'s value to $\text{sign}(S_{l,k})/\sqrt{d}$ where $S_{l,k} = \sum_{i \in \pi_{l,k}} g(\eps,\mu)_i \thetahat_i$. For a datapoint $z \in C_l$, the \emph{block sign} $s_k \in \{-1, +1\}$ is the common value $z_i \sqrt{d}$ for all $i \in \pi_{l,k}$ (which is well-defined since $z$ is block-constant within category $C_l$). For correctness, we need $\text{sign}(S_{l,k}) = s_k$ (the training datapoint's block sign) for all $k$.

\begin{lemma}[Sign consistency]
\label{lem:sign-consistency}
Suppose $d \ge \Omega(\eps^4 n^4 / \delta^2)$ and $n \le 1/(13824\,\eps^2)$. For a training datapoint $z$ from category $C_l$ satisfying $\langle v(z), \thetahat \rangle \ge \eps/n$, we have $\text{sign}(S_{l,k}) = s_k$ for all $k \in [d^{1/4}]$ simultaneously, with probability $\ge 1 - O(\delta)$ over the random partition.
\end{lemma}

\begin{proof}[Proof sketch]
The correlation bound decomposes by blocks as \(\frac{1}{\sqrt{d}}\sum_{k=1}^{d^{1/4}} s_k S_{l,k} \ge \frac{\eps}{n} - |S_0| - |\text{$\mu$-term}| \ge \frac{\eps}{2n}\),
where $S_0$ is the contribution of the pinned coordinates and the $\mu$-term $\sum_{i} g(\eps,\mu)_i \mu_i \thetahat_i$ arises from splitting $z_i - \mu_i = s_k/\sqrt{d} - \mu_i$ in non-pinned blocks. The $\mu$-term is at most $3456\eps^3$ by Cauchy--Schwarz (Lemma~\ref{lem:reweighting-norm}), which is at most $\eps/(4n)$ exactly when $n \le 1/(13824\,\eps^2)$; the pinned contribution satisfies $|S_0| \le O(\eps^{5/2} d^{-3/16})$ (Lemma~\ref{lem:pi0-contribution}), below $\eps/(4n)$ at the theorem's dimension.

Thus the average signed block score is $\ge \eps d^{1/4}/(2n)$. Each block score has variance $\le 288^2 \eps^4 d^{-1/4}$ over the random partition (Lemma~\ref{lem:variance-block}). By Chebyshev with $\kappa = \sqrt{d^{1/4}/\delta}$, the per-block deviation is $\le 288\eps^2/\sqrt{\delta}$, dimension-free. This is smaller than the signal $\eps d^{1/4}/(2n)$ exactly when $d^{1/4} \ge 576\eps n/\sqrt{\delta}$, i.e., $d \ge \Omega(\eps^4 n^4/\delta^2)$. A union bound over all $d^{1/4}$ blocks completes the argument. The full proof appears in Appendix~\ref{app:chebyshev-detailed}.
\end{proof}

\subsection{Supporting Lemmas}

\begin{lemma}[Variance of block scores]
\label{lem:variance-block}
For any block $k \in [d^{1/4}]$, $\text{Var}(S_{l,k}) \le 288^2 \eps^4 d^{-1/4}$.
\end{lemma}

\begin{proof}
Each coordinate $i$ is assigned to block $k$ independently with probability $d^{-1/4}$. The variance is $\sum_i g(\eps,\mu)_i^2 \thetahat_i^2 d^{-1/4}(1 - d^{-1/4}) \le d^{-1/4} \cdot \|g(\eps,\mu)\|_\infty^2 \|\thetahat\|^2 \le d^{-1/4} \cdot 288^2 \eps^4$, using $\|\thetahat\| \le 1$ and Lemma~\ref{lem:reweighting-norm}. (The per-coordinate estimate $\thetahat_i^2 \le 1$ would give only $288^2\eps^4 d^{3/4}$, which grows with $d$; the norm bound is what makes the Step~3 comparison in Appendix~\ref{app:chebyshev-detailed} close.)
\end{proof}

\begin{lemma}[Bounded contribution of $\pi_{l,0}$]
\label{lem:pi0-contribution}
For a fixed category $l$, with probability $1 - \delta$ over the random partition, $|\pi_{l,0}| \le 2\eps d^{5/8}/\sqrt{\delta}$ and the pinned contribution $S_0 = \sum_{i \in \pi_{l,0}} g(\eps,\mu)_i \thetahat_i(z_i - \mu_i)$ satisfies
\[
|S_0| \;\le\; 288\eps^2 \cdot \frac{2}{\sqrt{d}} \cdot \sqrt{|\pi_{l,0}|} \;\le\; 576\sqrt{2}\,\eps^{5/2}\,\delta^{-1/4}\, d^{-3/16} \;=\; O(\eps^{5/2} d^{-3/16}),
\]
which decays with $d$.
\end{lemma}

\begin{proof}
$\E[|\pi_{l,0}|] = \eps\sqrt{d}$. Since each coordinate is included independently with probability $\eps/\sqrt{d}$, the variance is $\text{Var}(|\pi_{l,0}|) \le \eps\sqrt{d}$. By Chebyshev's inequality,
\[
\Prob\!\left[|\pi_{l,0}| > \eps\sqrt{d} + t\right] \le \frac{\eps\sqrt{d}}{t^2}.
\]
Setting $t = \eps\sqrt{d} \cdot d^{1/8}/\sqrt{\delta}$ (so $\eps\sqrt{d}/t^2 = \delta/(\eps\sqrt{d} \cdot d^{1/4}) \le \delta$ for $d$ large) gives $|\pi_{l,0}| \le \eps\sqrt{d} + \eps d^{5/8}/\sqrt{\delta} \le 2\eps d^{5/8}/\sqrt{\delta}$ with probability $1 - \delta$. For $S_0$, each summand satisfies $|g(\eps,\mu)_i| \le 288\eps^2$ (Lemma~\ref{lem:reweighting-norm}) and $|z_i - \mu_i| \le 2/\sqrt{d}$, so by Cauchy--Schwarz over $\pi_{l,0}$, using $\|\thetahat\| \le 1$,
\[
|S_0| \;\le\; 288\eps^2 \cdot \frac{2}{\sqrt{d}} \sum_{i \in \pi_{l,0}} |\thetahat_i| \;\le\; 288\eps^2 \cdot \frac{2}{\sqrt{d}} \cdot \sqrt{|\pi_{l,0}|}\,\|\thetahat\| \;\le\; 576\sqrt{2}\,\eps^{5/2}\,\delta^{-1/4}\,d^{-3/16}. \qedhere
\]
\end{proof}

\subsection{Category Disjointness}
\label{app:disjointness}

Definition~\ref{def:block-uniform} requires the categories to be disjoint. The random construction of Section~\ref{sec:construction} delivers this except with exponentially small probability; the intuition is that an independent random partition almost never leaves a two-signed vector block-constant. We first record a uniform Chernoff estimate for the pinned parts, which strengthens the per-category Chebyshev bound of Lemma~\ref{lem:pi0-contribution} to an exponentially small failure probability (as needed both below and in the proof of Corollary~\ref{cor:soundness}, where a bound holding simultaneously for all $N$ categories is required and a per-category $1-\delta$ guarantee would be vacuous after the union bound).

\begin{lemma}[Uniform pin-count bound]
\label{lem:pin-count}
$|\pi_{l,0}| \sim \mathrm{Binomial}(d, \eps/\sqrt{d})$ for each $l$, so by the multiplicative Chernoff bound, $\Prob[|\pi_{l,0}| > 2\eps\sqrt{d}] \le e^{-\eps\sqrt{d}/3}$, and
\[
\Prob\big[\exists\, l \in [N]:\ |\pi_{l,0}| > 2\eps\sqrt{d}\big] \le N e^{-\eps\sqrt{d}/3}.
\]
\end{lemma}

\begin{lemma}[Category disjointness]
\label{lem:disjointness}
Under the hypotheses of Theorem~\ref{thm:exponential} (in particular $d = \Omega(n^8)$, $n = \Omega(1/\eps^2)$, and $N = n^2/\delta$ with $\delta^{-1} \in \mathrm{poly}(n)$), with probability $1 - e^{-\Omega(\eps\sqrt{d})}$ over the partitions $\Pi_1, \ldots, \Pi_N$, the categories $C_1, \ldots, C_N$ are pairwise disjoint.
\end{lemma}

\begin{proof}
Condition on two events. First, every block of every partition has size in $[d^{3/4}/2,\, 2d^{3/4}]$: each $|\pi_{l,k}|$ is $\mathrm{Binomial}(d - |\pi_{l,0}|,\, d^{-1/4})$ with mean $(1 \pm o(1))d^{3/4}$, so by Chernoff and a union bound over the $N d^{1/4}$ blocks this fails with probability at most $2Nd^{1/4}e^{-\Omega(d^{3/4})}$. Second, $|\pi_{l,0}| \le 2\eps\sqrt{d}$ for every $l$, which fails with probability at most $Ne^{-\eps\sqrt{d}/3}$ (Lemma~\ref{lem:pin-count}).

Fix $l \ne l'$ and condition on all pinned parts $\{\pi_{m,0}\}_{m \in [N]}$ and all pin values (the pin values are a function of the pinned parts and $\mu$, and, jointly with $\Pi_l$'s bin assignment, they determine $C_l$), as well as on $\Pi_l$'s bin assignment; then fix $z \in C_l$. The only randomness used below is the bin assignment of $\Pi_{l'}$'s unpinned coordinates, which remains i.i.d.\ uniform and independent of everything conditioned on, and the counting bounds that follow hold uniformly over every conditioned realization on the two events above. Since the block signs of $z$ do not all coincide (the two such vectors are excluded from $C_l$), it has at least one block of each sign under $\Pi_l$, hence at least $m \ge d^{3/4}/2$ coordinates carrying its minority sign, and at least $(d - |\pi_{l,0}|)/2 \ge d/4$ coordinates carrying its majority sign. For $z \in C_{l'}$ it is necessary that every coordinate of $z$ is either pinned in $\pi_{l',0}$ (with matching pin value -- a requirement we do not even need, as pins only make overlap harder) or lies in a block of $\Pi_{l'}$ that is monochromatic under $z$. Since $|\pi_{l',0}| \le 2\eps\sqrt{d} \le m/2$ for $d$ large, at least $m/2 \ge d^{3/4}/4$ minority coordinates of $z$ are unpinned in $\Pi_{l'}$, and each must then lie in a $\Pi_{l'}$-block containing no majority coordinate of $z$. But at least $d/4 - 2\eps\sqrt{d} \ge d/8$ majority coordinates are unpinned in $\Pi_{l'}$ and are assigned to the $d^{1/4}$ bins independently and uniformly, so the probability that some fixed bin receives none of them is at most $(1 - d^{-1/4})^{d/8} \le e^{-d^{3/4}/8}$. By a union bound over bins, with probability at least $1 - d^{1/4}e^{-d^{3/4}/8}$ every bin of $\Pi_{l'}$ contains a majority coordinate of $z$; on that event any unpinned minority coordinate of $z$ sits in a non-monochromatic block, so $z \notin C_{l'}$.

A union bound over the at most $2^{d^{1/4}}$ elements of $C_l$ and the $N^2$ ordered pairs $(l, l')$ bounds the total overlap probability by
\[
N^2\, 2^{d^{1/4}}\, d^{1/4} e^{-d^{3/4}/8} \;+\; 2Nd^{1/4}e^{-\Omega(d^{3/4})} \;+\; Ne^{-\eps\sqrt{d}/3} \;=\; e^{-\Omega(\eps\sqrt{d})},
\]
where the last equality holds because $\eps\sqrt{d}$ dominates both $d^{1/4}$ and $\log N$ at the dimension of Theorem~\ref{thm:exponential} ($\eps\sqrt{d} \ge \eps d^{1/4}\cdot d^{1/4}$ with $\eps d^{1/4} = \Omega(n^{3/2})$ when $d = \Omega(n^8)$ and $n = \Omega(1/\eps^2)$).
\end{proof}

\begin{remark}[Deterministic trimming]
\label{rem:trimming}
Disjointness can also be enforced deterministically: redefine $C'_l := C_l \setminus \bigcup_{l' < l} C_{l'}$. The trimmed categories are disjoint by construction, and Algorithm~\ref{alg:ra2} is unaffected, since its candidate for category $l$ is built from $\Pi_l$ alone. By Lemma~\ref{lem:disjointness}, nothing is trimmed except with probability $e^{-\Omega(\eps\sqrt{d})}$, so all tail bounds carry over up to that additive term.
\end{remark}

\section{Deferred Proofs}
\label{app:proofs}

\subsection{Soundness Transfer to \texorpdfstring{$\Delta_\mu$}{Delta\_mu}}
\label{app:soundness-transfer}

\begin{proof}[Proof of Lemma~\ref{lem:category-tail}]
Let $\bar{C}_l$ denote the category with the two excluded vectors (those whose block signs all coincide) added back; each $z \in \bar{C}_l$ is parameterized by independent Rademachers $s_1, \ldots, s_{d^{1/4}}$ (the block signs), and the inner product becomes $\sum_k s_k c_{l,k}$ where $c_{l,k} = \frac{1}{\sqrt{d}}\sum_{i \in \pi_{l,k}} y_i$. By Cauchy--Schwarz: $\sum_k c_{l,k}^2 \le \frac{M_l}{d}\|y\|^2$. Applying Hoeffding's inequality to the sum of independent bounded terms $s_k c_{l,k}$ with $\sum_k c_{l,k}^2 \le M_l \|y\|^2/d$ bounds the tail under $\text{Uniform}(\bar{C}_l)$ by $\exp(-t^2 d/(2M_l\|y\|^2))$. Conditioning on avoiding the excluded pair inflates any event's probability by at most $(1 - 2^{1-d^{1/4}})^{-1} \le 2$ and leaves $\E[z \mid l]$ unchanged (the pair is sign-flip symmetric), giving the stated bound.
\end{proof}

\begin{lemma}[Center-shift bound]
\label{lem:center-shift}
Let $\mu \in \R^d$ be an arbitrary fixed centering vector. Fix $y \in \R^d$ independent of a fresh draw $Z_0 \sim \Delta_\mu$. If $|\langle y, \E[z \mid l] - \mu \rangle| \le \Delta_{\mathrm{shift}}$ for all $l \in [N]$, then for $t \ge \Delta_{\mathrm{shift}}$:
\[
\Prob[\langle y, Z_0 - \mu \rangle \ge t] \le 2\exp\!\left(-\frac{(t - \Delta_{\mathrm{shift}})^2 d}{2M \|y\|^2}\right),
\]
where $M = \max_l \max_k |\pi_{l,k}|$.
\end{lemma}

\begin{proof}
Condition on $l$ and decompose: $\langle y, Z_0 - \mu \rangle = \langle y, Z_0 - \E[z \mid l] \rangle + \langle y, \E[z \mid l] - \mu \rangle$. The second term is bounded by $\Delta_{\mathrm{shift}}$, so if the LHS $\ge t$, then $\langle y, Z_0 - \E[z \mid l] \rangle \ge t - \Delta_{\mathrm{shift}}$. Apply the category tail bound (Lemma~\ref{lem:category-tail}) conditionally and average over $l$.
\end{proof}

\begin{proof}[Proof of Corollary~\ref{cor:soundness}]
Set $y = A\thetahat$ where $A = \text{diag}(g(\eps,\mu))$. Since $\|A\|_2 \le 288\eps^2$ and $\|\thetahat\| \le 1$, we have $\|y\| \le 288 \eps^2$.

We bound the shift $\Delta_{\mathrm{shift}} = \max_l |\langle A\thetahat, \E[z \mid l] - \mu \rangle|$ by splitting the coordinates by pinned status. For $i \notin \pi_{l,0}$ we have $\E[z_i \mid l] = 0$, so these coordinates contribute at most $|\sum_{i \notin \pi_{l,0}} g_i \thetahat_i \mu_i| \le \|g \odot \mu\| \|\thetahat\| \le 3456\eps^3$ -- the same Cauchy--Schwarz estimate as the $\mu$-term in Step~2 of Appendix~\ref{app:chebyshev-detailed} -- which is at most $\eps/(4n)$ since $n \le 1/(13824\,\eps^2)$. For $i \in \pi_{l,0}$ we have $|\E[z_i \mid l] - \mu_i| = |B_{i,l}/\sqrt{d} - \mu_i| \le 2/\sqrt{d}$, so by Cauchy--Schwarz (as in Lemma~\ref{lem:pi0-contribution}) these coordinates contribute at most $288\eps^2 \cdot (2/\sqrt{d}) \cdot \sqrt{|\pi_{l,0}|}$. Since the bound must hold simultaneously for all $N$ categories, we use the uniform pin-count bound (Lemma~\ref{lem:pin-count}) rather than the per-category Chebyshev estimate: except with probability $N e^{-\eps\sqrt{d}/3}$ -- which is $O(\delta^2/n^2)$ at the theorem's dimension -- every category has $|\pi_{l,0}| \le 2\eps\sqrt{d}$, so the pinned contribution is at most $576\sqrt{2}\,\eps^{5/2} d^{-1/4}$, itself at most $\eps/(8n)$ at the theorem's dimension. Hence $\Delta_{\mathrm{shift}} \le \eps/(4n) + \eps/(8n)$, and at the threshold $t = \eps/(2n)$ the budget $t - \Delta_{\mathrm{shift}} \ge \eps/(8n)$ survives.

By Lemma~\ref{lem:center-shift} with $t = \eps/(2n)$, and a union bound over the $n$ fresh samples:
\[
\Prob[\exists i: \langle A\thetahat, Z_i - \mu \rangle \ge \eps/(2n)] \le 2n \exp\!\left(-\frac{(\eps/8n)^2 d}{2M \cdot 288^2 \eps^4}\right) = 2n \exp\!\left(-\frac{d}{2^7 \cdot 288^2\, M \eps^2 n^2}\right).
\]
With $d = \Omega(M \cdot n^2 \log(n/\delta))$, this is $\le O(\delta^2/n^2)$.
\end{proof}

\subsection{Mean-Alignment via Bit-Fitting}

\begin{proposition}[Mean-alignment error]
\label{prop:mean-alignment}
The per-coordinate mean error satisfies $|\E_{z \sim \Delta_\mu}[z_i] - \mu_i| \le 1/(N\sqrt{d})$.
\end{proposition}

\begin{proof}
$\E[z_i] = \frac{1}{N}\sum_{l \in K_i} B_{i,l}/\sqrt{d}$ (non-pinned categories contribute $0$ by symmetry). Bit-fitting minimizes the distance to $\mu_i$, and adjusting one bit changes the sum by $2/(N\sqrt{d})$, so the error is at most $1/(N\sqrt{d})$.
\end{proof}

\subsection{Complete Sign Consistency Derivation}
\label{app:chebyshev-detailed}

\begin{proof}[Full proof of Lemma~\ref{lem:sign-consistency}]
Let $z$ be a training datapoint from category $C_l$ with block signs $(s_1, \ldots, s_{d^{1/4}})$, satisfying $\langle v(z), \thetahat \rangle \ge \eps/n$.

\textbf{Step 1: Block decomposition.}
Expanding the inner product:
\begin{align*}
\langle v(z), \thetahat \rangle &= \underbrace{\sum_{i \in \pi_{l,0}} g_i(z_i - \mu_i)\thetahat_i}_{S_0} + \sum_{k=1}^{d^{1/4}} \sum_{i \in \pi_{l,k}} g_i\!\left(\frac{s_k}{\sqrt{d}} - \mu_i\right)\thetahat_i \\
&= S_0 + \sum_{k=1}^{d^{1/4}} \left(\frac{s_k}{\sqrt{d}} S_{l,k} - \sum_{i \in \pi_{l,k}} g_i \mu_i \thetahat_i\right),
\end{align*}
where $g_i = g(\eps,\mu)_i$ and $S_{l,k} = \sum_{i \in \pi_{l,k}} g_i \thetahat_i$.

\textbf{Step 2: Bounding error terms.}
The $\mu$-term is the residual bias that appears when replacing $z_i$ by the block sign $s_k/\sqrt{d}$ in non-pinned blocks: since $z_i - \mu_i = (s_k/\sqrt{d} - \mu_i)$, each block contributes an extra $-\sum_{i \in \pi_{l,k}} g_i \mu_i \thetahat_i$. Summing over all blocks gives the total $\mu$-term $\sum_i g_i \mu_i \thetahat_i$, which we bound by Cauchy--Schwarz: $|\sum_i g_i \mu_i \thetahat_i| \le \|g \odot \mu\| \|\thetahat\| \le \|g \odot \mu\| \le \sqrt{\sum_i (288\eps^2)^2 \cdot (12\eps)^2/d} = 288 \cdot 12\eps^3 = 3456\eps^3$. Since $n \le 1/(13824\,\eps^2)$, this is at most $\eps/(4n)$. (This is the one error term that does not vanish as $d$ grows, and it is what fixes the upper end of the regime $n = \Theta(1/\eps^2)$.)

The pinned term: by Lemma~\ref{lem:pi0-contribution}, with probability $1 - \delta$ over the partition, $|S_0| \le 288\eps^2 \cdot (2/\sqrt{d}) \cdot \sqrt{|\pi_{l,0}|} \le 576\sqrt{2}\,\eps^{5/2}\,\delta^{-1/4}\,d^{-3/16}$. As $\delta^{-1} \in \text{poly}(n)$, this decays with $d$ and is below $\eps/(4n)$ at the theorem's dimension (it is dominated by the $\max\{\eps^4 n^4/\delta^2,\; n^8\log^4(n/\delta)\}$ terms of $d_0$).

So:
\[
\frac{1}{\sqrt{d}}\sum_k s_k S_{l,k} \ge \frac{\eps}{n} - \frac{\eps}{4n} - \frac{\eps}{4n} = \frac{\eps}{2n}.
\]

\textbf{Step 3: Per-block concentration.}
The average signed block score is $\ge \eps\sqrt{d}/(2n \cdot d^{1/4}) = \eps d^{1/4}/(2n)$.

For the random partition, $S_{l,k} = \sum_{i \notin \pi_{l,0}} g_i \thetahat_i \indicator[i \in \pi_{l,k}]$ where each indicator is independent with $\Prob[i \in \pi_{l,k}] = d^{-1/4}$. By Lemma~\ref{lem:variance-block}, $\text{Var}(S_{l,k}) \le 288^2 \eps^4 d^{-1/4}$, so $\sigma_k \le 288\eps^2 d^{-1/8}$.

Chebyshev with $\kappa = \sqrt{d^{1/4}/\delta}$:
\[
\Prob[|S_{l,k} - \E[S_{l,k}]| \ge \kappa\sigma_k] \le \delta/d^{1/4}.
\]
The deviation $\kappa\sigma_k = 288\eps^2 d^{1/8} \cdot d^{-1/8}/\sqrt{\delta} = 288\eps^2/\sqrt{\delta}$ is dimension-free, and is smaller than the signal $\eps d^{1/4}/(2n)$ exactly when
\[
d^{1/4} \ge \frac{288\eps n}{\sqrt{\delta}} \cdot 2 = \frac{576\eps n}{\sqrt{\delta}},
\]
i.e., $d \ge \Omega(\eps^4 n^4/\delta^2)$ -- which is precisely the hypothesis of Lemma~\ref{lem:sign-consistency}. Under this condition, each $s_k S_{l,k} > 0$.

\textbf{Step 4: Union bound.}
By the union bound over all $d^{1/4}$ blocks, sign consistency holds simultaneously with probability $\ge 1 - \delta$.
\end{proof}

\subsection{Reweighting Norm Bound}

\begin{proof}[Proof of Lemma~\ref{lem:reweighting-norm}]
$g(\eps,\mu)_i^2 = ((144\eps^2 - d\mu_i^2)/(1 - d\mu_i^2))^2 \le (144\eps^2/(1 - 144\eps^2))^2 \le (288\eps^2)^2$, using $d\mu_i^2 \le 144\eps^2$ and $\eps \le 1/(12\sqrt{2})$. Sum over $J$ for the norm bound.
\end{proof}

\subsection{Complete Proof of Theorem~\ref{thm:exponential}}
\label{app:complete-proof}

\begin{proof}[Proof of Theorem~\ref{thm:exponential}]
We assemble the individual lemma results. Fix $\eps \le \eps_0 = 1/(12\sqrt{2})$, $\delta \le \eps$ with $\delta^{-1} \in \text{poly}(n)$, $n \ge \Omega(1/\eps^2)$, and $d \ge \Omega(\max\{n^8 \log^4(n/\delta),\; \eps^4 n^4/\delta^2,\; n^2 \log(n + \delta^{-1})\})$ (by Remark~\ref{rem:dimension}, the $M$-dependent soundness constraint resolves to the $n^8 \log^4$ term).

\textbf{Step 1: Distribution construction.}
The block-uniform distribution $\Delta_\mu$ is constructed as in Section~\ref{sec:construction}: partition sampling (Algorithm~\ref{alg:partition-sampling}), bit-fitting (Algorithm~\ref{alg:distribution-sampling}), and category definition yield $N = n^2/\delta$ categories, each of size $2^{d^{1/4}} - 2$ (on the event that every block is non-empty, which fails with probability at most $d^{1/4}e^{-\Omega(d^{3/4})}$ and is folded into the failure events below), pairwise disjoint except with probability $e^{-\Omega(\eps\sqrt{d})}$ (Lemma~\ref{lem:disjointness}). By Proposition~\ref{prop:mean-alignment}, the per-coordinate mean error satisfies $|\E_{z \sim \Delta_\mu}[z_i] - \mu_i| \le 1/(N\sqrt{d})$.

\textbf{Step 2: Correlation bound.}
By Lemma~\ref{lem:correlation}, under the uniform prior $\mu \sim U = \text{Uniform}([-12\eps/\sqrt{d}, 12\eps/\sqrt{d}]^d)$, the expected total reweighted correlation satisfies
\[
\E_{\mu, S_n}\!\left[\sum_{j=1}^n \langle v(z_j), \thetahat \rangle\right] \ge 6\eps - 4\delta \ge 2\eps,
\]
where the last inequality uses $\delta \le \eps$. The correlation bound is derived from the distribution-free excess risk guarantee of the $\eps$-learner. Because excess risk is bounded for \emph{all} distributions -- including $\Delta_\mu$, whose mean satisfies $\|\E_{\Delta_\mu}[z] - \mu\|_\infty \le 1/(N\sqrt{d})$ -- the accuracy-correlation relationship (Lemma~\ref{lem:app-accuracy-correlation}) applies to $\Delta_\mu$ as well, with an additional error of at most $\sqrt{d} \cdot 1/(N\sqrt{d}) = 1/N = O(\delta/n^2)$, which is negligible.

By Markov's inequality on the complement, with probability $\ge 1 - \delta$ over $(\mu, S_n)$, at least $\Omega(n)$ training datapoints satisfy $\langle v(z_j), \thetahat \rangle \ge \eps/n$.

\textbf{Step 3: Soundness (no false positives).}
For a fresh sample $Z_0 \sim \Delta_\mu$ not in the training set, Corollary~\ref{cor:soundness} (proved via the category tail bound, Lemma~\ref{lem:category-tail}, and the center-shift bound, Lemma~\ref{lem:center-shift}) gives
\[
\Prob[\langle g(\eps,\mu) \odot (Z_0 - \mu), \thetahat \rangle \ge \eps/n] \le \exp\!\left(-\Omega\!\left(\frac{d}{M \eps^2 n^2}\right)\right)
\]
(the corollary in fact proves this at the smaller threshold $\eps/(2n)$, which implies the bound at the test's threshold $\eps/n$). With $d = \Omega(M \cdot n^2 \log(n/\delta))$, this probability is at most $O(\delta^2/n^2)$. A union bound over $N = n^2/\delta$ categories gives a total false-positive probability of at most $N \cdot O(\delta^2/n^2) = O(\delta)$.

\textbf{Step 4: Exact reconstruction via sign consistency.}
For each training datapoint $z$ from category $C_l$ satisfying $\langle v(z), \thetahat \rangle \ge \eps/n$, Lemma~\ref{lem:sign-consistency} guarantees that $\text{sign}(S_{l,k}) = s_k$ for all blocks $k \in [d^{1/4}]$ simultaneously, with probability $\ge 1 - O(\delta)$ over the random partition. This requires $d \ge \Omega(\eps^4 n^4/\delta^2)$. Since pinned coordinates are recovered by construction ($z'_i = B_{i,l}/\sqrt{d} = z_i$ for $i \in \pi_{l,0}$), the algorithm reconstructs $z$ exactly.

\textbf{Step 5: Union bound and conclusion.}
Define the following events:
\begin{itemize}
\item $E_0$: the categories are pairwise disjoint (Step 1; fails with probability $e^{-\Omega(\eps\sqrt{d})}$ by Lemma~\ref{lem:disjointness}, absorbed into $O(\delta)$ at the theorem's dimension),
\item $E_1$: at least $\Omega(n)$ training datapoints have $\langle v(z_j), \thetahat \rangle \ge \eps/n$ (Step 2),
\item $E_2$: no false positives among the $N$ categories (Step 3),
\item $E_3$: sign consistency holds for all training datapoints with large correlation (Step 4).
\end{itemize}
Each event fails with probability at most $O(\delta)$. By a union bound, all four hold simultaneously with probability $\ge 1 - O(\delta)$. Conditioning on $E_0 \cap E_1 \cap E_2 \cap E_3$, Algorithm~\ref{alg:ra2} reconstructs $\Omega(n)$ training datapoints exactly and produces no false positives. The algorithm iterates over $N = n^2/\delta$ categories, performing $O(d)$ work per category, for a total runtime of $O(Nd) = O(n^2 d / \delta) = \text{poly}(n, d, \delta^{-1})$.
\end{proof}

\subsection{Correlation Bound}

\begin{proof}[Proof of Lemma~\ref{lem:correlation}]
By the accuracy-correlation relationship (Lemma~\ref{lem:app-accuracy-correlation}):
$\langle \E[\thetahat], \mu \rangle \ge \|\mu\| - \eps - 2\delta$.
Under the prior $U$, $\E_\mu[\|\mu\|] \ge 4\eps$ (Lemma~\ref{lem:app-uniform-norm}), so $\E_\mu[\langle \E[\thetahat], \mu \rangle] \ge 3\eps - 2\delta$.

The reweighting vector $v(z)_i = g(\eps,\mu)_i(z_i - \mu_i)$ amplifies the correlation because $\E[v(z)_i \thetahat_i] = g(\eps,\mu)_i \cdot \text{Cov}(z_i, \thetahat_i)$. Since $d\mu_i^2 \le 144\eps^2$ by the parameter constraint, we have
\[
g(\eps,\mu)_i = \frac{144\eps^2 - d\mu_i^2}{1 - d\mu_i^2} \ge 0,
\]
with $g(\eps,\mu)_i = 144\eps^2$ at $\mu_i = 0$ and $g(\eps,\mu)_i = 0$ at $d\mu_i^2 = 144\eps^2$. Although $g(\eps,\mu)_i$ may be smaller than $1$ for small $\eps$, the reweighting redistributes weight towards coordinates with small $|\mu_i|$ (where $g_i \approx 144\eps^2$) and away from coordinates with large $|\mu_i|$ (where $g_i \approx 0$). Following the argument in \citet[Lemma~6.3]{attias2024information}, summing the reweighted correlation over all coordinates and all $n$ training datapoints yields $\E[\sum_{j=1}^n \langle v(z_j), \thetahat \rangle] \ge 6\eps - 4\delta$.
\end{proof}

\section{Additional Technical Details}
\label{app:technical}

\subsection{Notation Summary}

\begin{table}[ht]
\centering
\caption{Summary of notation}
\label{tab:notation}
\begin{tabular}{@{}ll@{}}
\toprule
\textbf{Symbol} & \textbf{Meaning} \\
\midrule
$d$ & Dimension of parameter space \\
$n$ & Sample size \\
$N$ & Number of candidates (Definition~\ref{def:game}) and of categories; $N = n^2/\delta$ \\
$\cD_\theta$, $\Delta_\theta$ & Ambient family; hard distribution $\mathrm{Uniform}(\Sigma)$, $\Sigma \sim \cD_\theta^{\otimes N}$ (Definition~\ref{def:game}) \\
$\eps,\;\delta$ & Excess risk and confidence parameters \\
$\cZ$ & Sample domain ($\{\pm 1/\sqrt{d}\}^d$); uppercase $Z_i$ denotes a random datapoint \\
$\cD_\mu$ & Product distribution over $\cZ$ with $\E[z_i] = \mu_i$ \\
$\Delta_\mu$ & Block-uniform hard distribution (Definition~\ref{def:block-uniform}), not a product \\
$\thetahat$ & Learned parameter ($\cA_n(S_n)$) \\
$\pi_{l,0}$ & Mean-alignment part of partition $\Pi_l$ \\
$\pi_{l,k}$ & Block $k$ of partition $\Pi_l$ \\
$B_{i,l}$ & Pin value for coordinate $i$ in category $l$ \\
$g(\eps, \mu)$ & Reweighting vector \\
$v(z)$ & Weighted correlation vector \\
$\odot$ & Hadamard (entrywise) product \\
$S_{l,k}$ & Block score: $\sum_{i \in \pi_{l,k}} g_i \thetahat_i$ \\
\bottomrule
\end{tabular}
\end{table}

\subsection{Parameter Relationships}

The main theorems require: (1) $n = \Theta(1/\eps^2)$, i.e.\ $\Omega(1/\eps^2) \le n \le 1/(13824\,\eps^2)$, the regime of Theorem~\ref{thm:exponential}: the lower end is the sample-complexity threshold inherited from the membership-inference lower bound of \citet{attias2024information}, and the upper end is fixed by the $\mu$-term budget in the sign-consistency and soundness steps (Appendix~\ref{app:chebyshev-detailed}; Corollary~\ref{cor:soundness}); (2) $d \ge \max\{\Omega(n^8 \log^4(n/\delta)),\; \Omega(\eps^4 n^4/\delta^2),\; \Omega(n^2 \log(n + \delta^{-1}))\}$ (the first term resolves the self-referential block-size constraint of Remark~\ref{rem:dimension} below: each non-pinned coordinate is assigned uniformly to one of $d^{1/4}$ blocks, so the expected block size is $(d - |\pi_{l,0}|)/d^{1/4} \approx d^{3/4}$ and standard concentration gives $M = O(d^{3/4})$ with high probability; substituting into $d = \Omega(M \cdot n^2 \log(n/\delta))$ yields $d^{1/4} = \Omega(n^2 \log(n/\delta))$, i.e.\ $d = \Omega(n^8 \log^4(n/\delta))$); (3) $\delta \le \eps$, $\delta^{-1} \in \text{poly}(n)$; (4) $N = n^2/\delta$; (5) $\mu \in [-12\eps/\sqrt{d}, 12\eps/\sqrt{d}]^d$.

\begin{remark}[Resolving the dimension requirement]
\label{rem:dimension}
The dimension requirement in Theorem~\ref{thm:exponential} is the maximum of three independent constraints: soundness ($d = \Omega(n^8 \log^4(n/\delta))$: the requirement $d = \Omega(M \cdot n^2 \log(n/\delta))$ of Corollary~\ref{cor:soundness} is self-referential, since the maximum block size $M = \max_l \max_k |\pi_{l,k}|$ depends on $d$, and it resolves via $M = O(d^{3/4})$ with high probability; see item~(2) above), sign consistency ($d = \Omega(\eps^4 n^4/\delta^2)$, Lemma~\ref{lem:sign-consistency}), and mean alignment ($d = \Omega(n^2 \log(n + \delta^{-1}))$). Under the theorem's hypotheses with $\delta^{-1} = O(n^3)$ -- in particular for any $\delta^{-1}$ of low polynomial degree -- soundness dominates; for polynomially smaller $\delta$ the sign-consistency term $\eps^4 n^4/\delta^2$ takes over. All three are absorbed by the single $\Omega(\cdot)$ in the theorem statement.
\end{remark}

\subsection{Supplementary Lemmas}

\begin{lemma}[Expected norm of uniform prior]
\label{lem:app-uniform-norm}
For $U = \text{Uniform}([-12\eps/\sqrt{d}, 12\eps/\sqrt{d}]^d)$, $\E_{\mu \sim U}[\|\mu\|] \ge 4\eps$.
\end{lemma}

\begin{proof}
$\E[\|\mu\|^2] = d \cdot 144\eps^2/(3d) = 48\eps^2$, and $\|\mu\| \le \sqrt{d} \cdot 12\eps/\sqrt{d} = 12\eps$ almost surely, so $\|\mu\|^2 \le 12\eps\,\|\mu\|$ and $\E[\|\mu\|] \ge \E[\|\mu\|^2]/(12\eps) = 4\eps$.
\end{proof}

\begin{lemma}[Accuracy-correlation relationship]
\label{lem:app-accuracy-correlation}
For an $\eps$-learner and $f(\theta,z) = -\langle \theta, z \rangle$: $\langle \E[\thetahat], \mu \rangle \ge \|\mu\| - \eps - 2\delta$.
\end{lemma}

\begin{proof}
The optimal $\theta^* = \mu/\|\mu\|$ achieves risk $-\|\mu\|$. Excess risk $\le \eps$ gives $\langle \thetahat, \mu \rangle \ge \|\mu\| - \eps$ with probability at least $1 - \delta$ over $S_n \sim \cD_\mu^{\otimes n}$. Decomposing the expectation over the high-probability event and its complement:
\begin{align*}
\langle \E[\thetahat], \mu \rangle
&= \E[\langle \thetahat, \mu \rangle] \\
&= \Prob[E] \cdot \E[\langle \thetahat, \mu \rangle \mid E] + \Prob[E^c] \cdot \E[\langle \thetahat, \mu \rangle \mid E^c] \\
&\ge (1 - \delta)(\|\mu\| - \eps) + \delta \cdot (-\|\mu\|) \\
&= \|\mu\| - \eps - 2\delta\|\mu\| \\
&\ge \|\mu\| - \eps - 2\delta,
\end{align*}
where $E$ denotes the event that excess risk $\le \eps$ holds, the third line uses $\langle \thetahat, \mu \rangle \ge -\|\thetahat\|\|\mu\| \ge -\|\mu\|$ (since $\|\thetahat\| \le 1$), and the last step uses $\|\mu\| \le 1$.
\end{proof}

\section{Generalizations, Barriers, and Open Directions}
\label{app:discussion}

This appendix collects the discussion of the adversarial model and of the reach of the framework: which parts of the model are necessary, which relax, and where the bottlenecks to generalization lie. Appendix~\ref{app:sampling} shows that the adversary's knowledge of the distribution parameter relaxes to sampling access at polynomial cost. Appendix~\ref{app:barrier} shows that the remaining feature of the model -- a hard distribution chosen per learner -- is necessary at the standard sample size. Appendix~\ref{app:axioms} isolates the properties of the category construction that the exponential-domain argument actually uses. Appendices~\ref{app:continuous} and~\ref{app:exp-families} discuss continuous distributions and the exponential-family extension we regard as the main open problem.

\subsection{Sampling Access Suffices}
\label{app:sampling}

The adversary of Theorem~\ref{thm:exponential} receives the realized mean $\mu$. This is a convenience, not a load-bearing assumption: it relaxes to sampling access at the realized parameter -- the resource practical membership inference attacks assume \citep{shokri2017membership, carlini2022membership} -- at polynomial cost. The game is otherwise unchanged: the family, the prior, and the partition-sampling and bit-fitting rules stay fixed and public, and the partitions $\Pi_l$ are the adversary's own coins; only $\mu$ is withheld.

\begin{proposition}[Sampling access suffices]
\label{prop:sampling}
Consider the setting of Theorem~\ref{thm:exponential}, with the game modified so that the realized $\mu$ is withheld from the adversary, which instead receives $m_1$ fresh i.i.d.\ samples from the product distribution $\cD_\mu$ and $m_2$ fresh i.i.d.\ samples from $\Delta_\mu$, drawn independently of the training set. There is an absolute constant $C$ such that whenever
\[
m_1 \;\ge\; C\, n^2 d^{1/4} \log(d/\delta)
\qquad\text{and}\qquad
m_2 \;\ge\; C\, N \log(N/\delta),
\]
the conclusion of Theorem~\ref{thm:exponential} continues to hold with probability $1 - O(\delta)$ for the adversary that runs Algorithm~\ref{alg:ra2} with the projected empirical estimate $\hat{\mu}$ in place of $\mu$, the pin values read off the $\Delta_\mu$-samples, and acceptance threshold $3\eps/(4n)$ in place of $\eps/n$. The total budget is thus $m = \Omega(n^2 d^{1/4} + N \log N)$ up to the logarithmic factors -- polynomial in $n$, $d$, and $\delta^{-1}$ -- and we have made no attempt to optimize it.
\end{proposition}

\begin{proof}
Throughout, condition on the events of the original analysis (correlation, soundness, sign consistency, disjointness), which are unaffected by the extra samples. Write $\Delta\mu = \hat{\mu} - \mu$, $g = g(\eps,\mu)$, $\hat{g} = g(\eps,\hat{\mu})$, and $\Delta g = \hat{g} - g$.

\textbf{Step 1: Estimation and projection.}
Let $\hat{\mu}$ be the coordinate-wise empirical mean of the $m_1$ samples from $\cD_\mu$, projected onto the box $[-12\eps/\sqrt{d}, 12\eps/\sqrt{d}]^d$. Since $\mu$ lies in the box, the projection can only decrease each $|\Delta\mu_i|$, and it keeps the bounds of Lemma~\ref{lem:reweighting-norm} valid for $\hat{g}$. Each coordinate satisfies $z_i \in \{\pm 1/\sqrt{d}\}$, so by Hoeffding's inequality and a union bound over the $d$ coordinates, with probability $1 - \delta$,
\[
|\Delta\mu_i| \;\le\; \tau := \sqrt{\frac{2\log(4d/\delta)}{m_1 d}} \quad \text{for every } i \in [d],
\]
hence $\|\Delta\mu_J\|_2 \le \sqrt{|J|}\,\tau$ for every $J \subseteq [d]$.

\textbf{Step 2: Perturbation of the statistic.}
Differentiating $g_i = (144\eps^2 - d\mu_i^2)/(1 - d\mu_i^2)$ under the constraint $d\mu_i^2 \le 144\eps^2$ gives $|\partial g_i/\partial \mu_i| \le 2d\,|\mu_i|\,(1 - 144\eps^2)^{-2} \le 96\eps\sqrt{d}$ for $\eps \le \eps_0$, so $|\Delta g_i| \le 96\eps\sqrt{d}\,|\Delta\mu_i|$. For any $x \in \cZ$ (so $|x_i - \hat{\mu}_i| \le 2/\sqrt{d}$ after projection), the adversary's statistic decomposes as
\[
\hat{g} \odot (x - \hat{\mu}) - g \odot (x - \mu) \;=\; \Delta g \odot (x - \hat{\mu}) \;-\; g \odot \Delta\mu,
\]
whose two parts contribute, after pairing with $\|\thetahat\| \le 1$,
\[
|\langle \Delta g \odot (x - \hat{\mu}), \thetahat\rangle| \le \tfrac{2}{\sqrt{d}}\|\Delta g\|_2 \le 192\eps\|\Delta\mu\|_2,
\qquad
|\langle g \odot \Delta\mu, \thetahat\rangle| \le \|g\|_\infty \|\Delta\mu\|_2 \le 288\eps^2\|\Delta\mu\|_2.
\]
These are two additive parts of one and the same difference, so they share a single $\eps/(8n)$ estimation allotment, half each: it suffices that $\|\Delta\mu\|_2 \le 1/(3072\,n)$ and $\|\Delta\mu\|_2 \le 1/(4608\,\eps n)$, i.e.\ (via $\|\Delta\mu\|_2 \le \sqrt{d}\,\tau$) that $m_1 = \Omega(n^2\log(d/\delta))$ and $m_1 = \Omega(\eps^2 n^2 \log(d/\delta))$ respectively -- both dominated by the block-score constraint of Step~3. Hence the statistic of every candidate moves by at most $\eps/(8n)$.

\textbf{Step 3: Perturbation of the block scores (the binding constraint).}
The decoder reads $\text{sign}(\hat{S}_{l,k})$ with $\hat{S}_{l,k} = \sum_{i \in \pi_{l,k}} \hat{g}_i \thetahat_i$. By Cauchy--Schwarz and Step~1,
\begin{align*}
|\hat{S}_{l,k} - S_{l,k}| \;&\le\; \|\Delta g_{\pi_{l,k}}\|_2 \,\|\thetahat\| \;\le\; 96\eps\sqrt{d}\cdot\sqrt{|\pi_{l,k}|}\;\tau \\
&\le\; 96\eps\sqrt{d}\cdot\sqrt{2d^{3/4}}\;\tau \;=\; O\!\left(\eps\, d^{3/8}\sqrt{\tfrac{\log(d/\delta)}{m_1}}\right),
\end{align*}
using the block-size bound $|\pi_{l,k}| \le 2d^{3/4}$ from the proof of Lemma~\ref{lem:disjointness}. The signal is $\eps d^{1/4}/(2n)$ (Step~3 of Appendix~\ref{app:chebyshev-detailed}); requiring the perturbation to be at most $\eps d^{1/4}/(8n)$ forces $m_1 = \Omega(n^2 d^{1/4}\log(d/\delta))$ -- the binding constraint, and the one that grows with $d$. Sign consistency then survives: the Chebyshev deviation $288\eps^2/\sqrt{\delta}$ plus the perturbation $\eps d^{1/4}/(8n)$ stay below the signal exactly when $d^{1/4} \ge 768\eps n/\sqrt{\delta}$, the same asymptotics as Lemma~\ref{lem:sign-consistency}.

\textbf{Step 4: Recall and soundness at the shifted threshold.}
By Step~2, every candidate's statistic moves by at most $\eps/(8n)$. A training point flagged by the exact analysis has true statistic $\ge \eps/n$, hence perturbed statistic $\ge \eps/n - \eps/(8n) \ge 3\eps/(4n)$: it passes the modified threshold. Conversely, a fresh point passing the modified threshold has true statistic $\ge 3\eps/(4n) - \eps/(8n) = 5\eps/(8n) \ge \eps/(2n)$, so the tail of Corollary~\ref{cor:soundness} -- proven at threshold $\eps/(2n)$ -- applies verbatim and the false-positive analysis of Appendix~\ref{app:complete-proof} goes through unchanged. This is why the relaxation runs in the same regime $n \le 1/(13824\,\eps^2)$ as the theorem: the detection margin $\eps/(2n)$ and the soundness slack are already budgeted there, and estimation consumes only the residual $\eps/(8n)$.

\textbf{Step 5: Pins by coverage.}
The pin values are the one discrete function of $\mu$, and the adversary does not recompute them from $\hat{\mu}$: it reads them off the fresh $\Delta_\mu$-samples. Each such sample belongs to its category $l$ and is block-constant on the non-pinned part of $\Pi_l$; by the argument of Lemma~\ref{lem:disjointness} (which never uses the pin values), it is block-constant on no other partition, except with probability $e^{-\Omega(\eps\sqrt{d})}$. The adversary therefore classifies each sample by checking block-constancy against its own partitions, and within a category the pinned coordinates are constant across samples, so a single sample of category $l$ reveals all of $(B_{i,l})_{i \in \pi_{l,0}}$. Since $\Delta_\mu$ draws $l \sim \text{Uniform}([N])$, covering the categories is a coupon-collector problem: $m_2 \ge C N\log(N/\delta)$ covers every category with probability $1 - \delta$. ($m_2 = \Omega(N)$ already covers a constant fraction of the categories, which preserves the $\Omega(n)$ yield since the training points' categories are uniform as well; we state the full-coverage version for simplicity.)

A union bound over the failure events of Steps~1--5 and of the original analysis completes the proof.
\end{proof}

\subsection{Necessity of a Per-Learner Hard Distribution}
\label{app:barrier}

The remaining feature of the adversarial model is that the hard distribution is chosen per learner. This is not a modeling preference but a necessity at the standard sample size (Proposition~\ref{prop:barrier}): against any \emph{single}, publicly known distribution, every $\eps$-learner can be immunized on it -- and on a whole neighborhood of it, not a knife-edge -- by a hybrid that outputs a fixed vector whenever a goodness-of-fit test passes. Distribution-free accuracy is exactly the property that forces the learner to encode its sample in $\thetahat$; hardness must therefore be exhibited by \emph{some} distribution, chosen per learner.

\begin{proposition}[Immunizing hybrid]
\label{prop:barrier}
Let $\cD$ be any fixed, publicly known distribution over $\cZ = \{\pm 1/\sqrt{d}\}^d$ (so $\|z\| = 1$ for all $z$), with mean $\nu = \E_{\cD}[z]$ and population minimizer $\theta^*(\cD)$ of the linear loss ($\theta^*(\cD) = \nu/\|\nu\|$ if $\nu \ne 0$, and any fixed $\theta_0 \in \Theta$ otherwise). Fix $\delta \in (0,1)$, let $c = 1 + \sqrt{2\log(1/\delta)}$, and let $n \ge 9c^2/\eps^2$. Given any $\eps$-learner $\cA$, define the hybrid
\[
\cA'(S) \;=\;
\begin{cases}
\theta^*(\cD) & \text{if } \|\bar{z} - \nu\| \le 2c/\sqrt{n},\\[2pt]
\cA(S) & \text{otherwise},
\end{cases}
\qquad \bar{z} = \frac{1}{n}\sum_{i=1}^n z_i.
\]
Then:
\emph{(a)} on every distribution over $\cZ$, $\cA'$ has excess risk at most $4\eps$ with probability at least $1 - 2\delta$ -- an $\eps$-learner after rescaling $\eps$ and $\delta$ by constants; and
\emph{(b)} on every distribution $\cD'$ over $\cZ$ whose mean $\nu'$ satisfies $\|\nu' - \nu\| \le c/\sqrt{n}$ -- in particular on $\cD$ itself -- with probability at least $1 - \delta$, $\cA'$ outputs the fixed vector $\theta^*(\cD)$, so its output carries no information about the training set beyond the single bit that the test passed, and no reconstruction from it is possible.
\end{proposition}

\begin{proof}
\emph{Excess risk of the fixed vector, exactly.} For a distribution $\cD'$ over $\cZ$ with mean $\nu'$, the population risk of the linear loss is $F_{\cD'}(\theta) = -\langle\theta, \nu'\rangle$, minimized over $\Theta$ at value $-\|\nu'\|$. Writing $\Delta = \|\nu' - \nu\|$, for $\nu \ne 0$,
\[
\mathrm{exc}\big(\theta^*(\cD); \cD'\big) \;=\; \|\nu'\| - \Big\langle \frac{\nu}{\|\nu\|}, \nu'\Big\rangle \;=\; \frac{\Delta^2 - (\|\nu'\| - \|\nu\|)^2}{2\|\nu\|} \;\le\; \min\Big\{2\|\nu'\|,\; \frac{\Delta^2}{2\|\nu\|}\Big\},
\]
and this holds at \emph{every} gap $\Delta$, not only inside the test radius. The second branch is second order in the gap, giving the $\eps/2$ bound below when $\|\nu\| \ge \eps$; the first branch covers small $\|\nu\|$, and also the degenerate case $\nu = 0$, where the excess risk of any fixed $\theta_0$ is trivially at most $2\|\nu'\|$. So the argument below applies to every fixed $\cD$.

\emph{Tail for the test statistic.} Every point of $\cZ$ has unit norm, so for any distribution $\cD'$ with mean $\nu'$, the map $S \mapsto \|\bar{z} - \nu'\|$ has bounded differences $2/n$ and expectation at most $\sqrt{\E\|\bar{z} - \nu'\|^2} \le 1/\sqrt{n}$. McDiarmid's inequality gives, for every $s \ge 0$ and every distribution on $\cZ$,
\[
\Prob\big[\|\bar{z} - \nu'\| \ge (1+s)/\sqrt{n}\big] \le e^{-s^2/2},
\]
so with $c = 1 + \sqrt{2\log(1/\delta)}$: $\Prob[\|\bar{z} - \nu'\| \ge c/\sqrt{n}] \le \delta$. The hypothesis $n \ge 9c^2/\eps^2$ gives $3c/\sqrt{n} \le \eps$.

\emph{(a) Accuracy.} Let the distribution at hand have mean $\nu'$ and gap $\Delta$. If $\Delta > 3c/\sqrt{n}$: passing the test forces $\|\bar{z} - \nu'\| \ge \Delta - 2c/\sqrt{n} > c/\sqrt{n}$, an event of probability at most $\delta$; so with probability $\ge 1 - \delta$ the test fails, $\cA$ runs, and $\cA$'s own $(\eps,\delta)$-guarantee applies -- total failure probability at most $2\delta$. If $\Delta \le 3c/\sqrt{n} \le \eps$: whenever the test passes, the display above bounds the excess risk \emph{deterministically} -- by $\Delta^2/(2\|\nu\|) \le \eps^2/(2\eps) = \eps/2$ if $\|\nu\| \ge \eps$, and by $2\|\nu'\| \le 2(\|\nu\| + \Delta) < 4\eps$ if $\|\nu\| < \eps$ -- while whenever it fails, $\cA$'s guarantee applies. In all cases the excess risk is at most $4\eps$ with probability at least $1 - 2\delta$.

\emph{(b) Immunity on a neighborhood.} Let $\cD'$ have mean $\nu'$ with $\Delta = \|\nu' - \nu\| \le c/\sqrt{n}$. By the triangle inequality, $\|\bar{z} - \nu\| \le \|\bar{z} - \nu'\| + c/\sqrt{n}$, so the test passes whenever $\|\bar{z} - \nu'\| \le c/\sqrt{n}$, an event of probability at least $1 - \delta$ by the tail bound. On it the output is $\theta^*(\cD)$, a vector fixed before the sample was drawn; by data processing, nothing about $S$ is recoverable from it beyond the single bit that the test passed. The case $\cD' = \cD$ is $\Delta = 0$.
\end{proof}

The hypothesis $n \ge 9c^2/\eps^2 = \Omega(\log(1/\delta)/\eps^2)$ is the standard sample size for excess risk $\eps$ at confidence $\delta$, and it lies above the window $n \le 1/(13824\,\eps^2)$ of Theorem~\ref{thm:exponential}. The proposition is a statement about the adversarial model (at that sample size, no theorem with a single fixed hard distribution is possible), not about that window. Two observations mitigate the per-learner choice, and both are true of our construction:
\begin{enumerate}[itemsep=1pt, topsep=2pt]
    \item \emph{The choice is minimal and almost learner-oblivious.} The family, the prior $\mu \sim \mathrm{Uniform}([-12\eps/\sqrt{d}, 12\eps/\sqrt{d}]^d)$, and the partition-sampling and bit-fitting rules are fixed with no reference to the learner; only the realized $\mu$ (and through it the pin values) depends on it, extracted by an averaging argument over that fixed prior (Lemma~\ref{lem:correlation}), as in \citet{attias2024information}.
    \item \emph{Equivalently, a statement about a fixed class.} Theorem~\ref{thm:exponential} reads as: no $\eps$-learner resists reconstruction on every block-uniform distribution in this family. Under this reading the adversary ``chooses'' nothing.
\end{enumerate}

\subsection{Decodable Category Families}
\label{app:axioms}

The block-uniform construction of Section~\ref{sec:exponential} is one member of a broader class of families to which the same argument applies.

\begin{remark}[Decodable category families]
\label{rem:axioms}
Besides the correlation bound (Lemma~\ref{lem:correlation}) and $N = n^2/\delta$ disjoint categories (Definition~\ref{def:block-uniform}; the collision step of Theorem~\ref{thm:reduction}), the exponential-domain argument uses four properties of the constructed family:
\begin{enumerate}[label=(\roman*), itemsep=1pt, topsep=2pt]
    \item \emph{polynomially many categories};
    \item \emph{per-category decodability}: from the learner's output $\thetahat$, a single candidate per category is computable in polynomial time and, with probability $1-O(\delta)$, equals every training point of that category that clears the test threshold (here: block-constancy plus sign decoding, Algorithm~\ref{alg:ra2} and Lemma~\ref{lem:sign-consistency});
    \item \emph{a sub-Gaussian category tail}: for the relevant test vectors $y$, the statistic $\langle y, z - \E[z \mid \text{category}]\rangle$ has a sub-Gaussian tail (here Lemma~\ref{lem:category-tail}, replacing the product-measure tail bound of \citet{attias2024information}, proven from the block-sign representation alone);
    \item \emph{mean alignment}: the category means are aligned with the target parameter, in two senses -- $\max_l |\langle y, \E[z \mid l] - \mu\rangle|$ lies below the test threshold for the relevant test vectors (in our construction via the center-shift bound, Lemma~\ref{lem:center-shift}, as established in the proof of Corollary~\ref{cor:soundness}), and $\E_{\Delta}[z]$ is close enough to $\mu$ that the accuracy-to-correlation transfer costs less than the detection margin (in our construction, the role of the pinned coordinates and the bit-fitting rule).
\end{enumerate}
Any ``decodable category family'' satisfying (i)--(iv) supports the proof of Theorem~\ref{thm:exponential}; uniformity within categories is a convenience, not a necessity -- biased within-category distributions with bounded sub-Gaussian norm satisfy (iii) with adjusted constants. We state this as the interface for future instantiations.
\end{remark}

\subsection{Approximate Reconstruction for Continuous Distributions}
\label{app:continuous}

Under a continuous data distribution, outputting a training point exactly is a measure-zero event, so the reconstruction goal itself must change: the standard notion is \emph{approximate} reconstruction -- output $z'$ with $\|z' - z\| \le \rho$ for some training point $z$ \citep{balle2022reconstructing}. The reduction template extends to this goal. Run Algorithm~\ref{alg:ra1} over a $\rho$-net of the support in place of the sampled candidate set, and rely on the test having a \emph{margin}, so that a training point is still detected through its nearest net point. The false-positive union bound then runs over the net size $C$ rather than $|\Sigma|$: whenever the test's false-positive rate is $o(\delta/C)$, the argument of Theorem~\ref{thm:reduction} recovers $\Omega(n \cdot q(n))$ training points up to error $\rho$ -- for instance, for bounded-support distributions in fixed dimension, where a polynomial-size net exists. Our SCO test does have such a margin: the statistic $x \mapsto \langle g(\eps,\mu) \odot (x - \mu), \thetahat\rangle$ is Lipschitz in $x$ with constant $\|g\|_\infty\|\thetahat\| \le 288\eps^2$, while soundness holds already at threshold $\eps/(2n)$ when $n = \Theta(1/\eps^2)$ (Corollary~\ref{cor:soundness}); testing net points at threshold $\eps/(2n)$ instead of $\eps/n$, a training point with statistic at least $\eps/n$ is detected through its nearest net point whenever $288\eps^2\rho \le \eps/(2n)$, i.e.\ $\rho \le 1/(576\,\eps n)$. This is the same slack that Appendix~\ref{app:sampling} spends on estimation, so combining the two splits it.

We state the caveat explicitly: \emph{no end-to-end continuous theorem is claimed}. A $\rho$-net of polynomial size requires essentially fixed dimension, while our test's margin is proven in the high-dimensional regime of Theorem~\ref{thm:exponential} ($d = \Omega(n^8\log^4(n/\delta))$), so the two ingredients do not currently meet: an end-to-end continuous theorem needs a membership test at the required rates whose margin lives over a polynomially sized net. What the mechanism does show is that continuity is no obstacle to the reduction template itself; we regard closing this gap as a concrete open direction, and the exponential-family route below as the most promising path toward it.

\subsection{Beyond Product Bernoulli: Exponential Families}
\label{app:exp-families}

The sole component of our chain tied to the product Bernoulli family is the correlation bound (Lemma~\ref{lem:correlation}, following Lemma~6.3 of \citealp{attias2024information}), which converts $\eps$-accuracy into per-sample signal and rests on a fingerprinting lemma for that family \citep{bun2014fingerprinting, steinke2017tight}. The fingerprinting method is no longer confined there: \citet{kamath2022new} prove a fingerprinting lemma for arbitrary exponential families $\mathcal{E}(T, h)$ (their Lemma~3.2), with the natural parameter drawn uniformly from a box and the correlation measured against the sufficient statistics $T(X_i)$; no privacy assumption enters that step, no coordinate-independence of the samples is required, and the binary-product and Gaussian lemmas are recovered as special cases.

We therefore state as our main open problem: \emph{establish a membership inference impossibility at the rates the reduction consumes -- recall at least $q(n)$ and false-positive rate $O(\delta^2/n^2)$ under the hard distribution of Definition~\ref{def:game} (Definition~\ref{def:detection-rate}) -- for general exponential families.} Three steps appear to be needed: (1) re-parameterize the SCO instance so that excess risk controls the estimation error of the natural parameter $\eta$ (a natural candidate is the negative log-likelihood $f(\theta, x) = Z(\theta) - \langle\theta, T(x)\rangle$, convex because the log-partition function $Z$ is); (2) read the per-sample signal off $T(x)$ rather than $x$ -- the two coincide in our instance, where $T(x) = x$, which is why the current proof need not distinguish them -- making the reconstruction target a sample through its sufficient statistics; and (3) supply a concentration bound for the new family in place of Lemma~\ref{lem:category-tail}. Since the reduction (Theorem~\ref{thm:reduction}) is distribution-agnostic, any such result upgrades to reconstruction with no further work, covering continuous families such as Gaussians and removing the Bernoulli assumption rather than re-proving it family by family.

\end{document}